\documentclass[10pt,journal,compsoc]{IEEEtran}
\usepackage{booktabs} 
\usepackage{multirow}
\usepackage{graphicx}
\usepackage{subfigure}
\usepackage{enumitem}
\usepackage{multirow}
\usepackage{amssymb}
\usepackage{amsmath}
\usepackage{xurl}
\usepackage{bbm} 
\usepackage{xcolor}

\ifCLASSOPTIONcompsoc
  \usepackage[nocompress]{cite}
\else
  \usepackage{cite}
\fi
\ifCLASSINFOpdf
\else
\fi
\begin{document}
%
\title{REPLAY: Modeling Time-Varying Temporal Regularities of Human Mobility for Location Prediction over Sparse Trajectories}
%
%
%
%

\author{Bangchao~Deng,~
        Bingqing~Qu,~
        Pengyang~Wang,~
        Dingqi~Yang*,~
        Benjamin~Fankhauser,~
        and~Philippe~Cudre-Mauroux
\IEEEcompsocitemizethanks{\IEEEcompsocthanksitem Bangchao Deng, Pengyang Wang, and Dingqi Yang are with the State Key Laboratory of Internet of Things for Smart City and Department of Computer and Information Science, University of Macau, Macao SAR, China, E-mail: yc37980@um.edu.mo, pywang@um.edu.mo, dingqiyang@um.edu.mo. Bingqing Qu is with BNU-HKBU United International College, China, E-mail: bingqingqu@uic.edu.cn. Benjamin Fankhauser is with Bern University of Applied Sciences, Switzerland, E-mail: benjamin.fankhauser@bfh.ch. Philippe Cudre-Mauroux is with the University of Fribourg, Switzerland, E-mail: philippe.cudre-mauroux@unifr.ch. \protect\\
\IEEEcompsocthanksitem *Corresponding author: Dingqi Yang (email: dingqiyang@um.edu.mo)}
\thanks{Manuscript received April 19, 2005; revised August 26, 2015.}}

%
%

\markboth{Journal of \LaTeX\ Class Files,~Vol.~14, No.~8, August~2015}%
{Shell \MakeLowercase{\textit{et al.}}: Bare Demo of IEEEtran.cls for Computer Society Journals}
%



\IEEEtitleabstractindextext{%
\begin{abstract}
    Next-location prediction aims to forecast which location a user is most likely to visit given the user's historical data. As a sequence modeling problem by nature, it has been widely addressed using Recurrent Neural Networks (RNNs). To tackle the intrinsic sparsity issue of real-world user mobility traces, spatiotemporal contexts have been shown as significantly useful. Existing solutions mostly incorporate spatiotemporal distances between locations in mobility traces, either by integrating them into the RNN units as additional information, or utilizing them to search for informative historical hidden states to improve prediction. However, such distance-based methods fail to capture the time-varying temporal regularities of human mobility, where human mobility is often more regular in the morning than in other time periods, for example; this suggests the usefulness of the actual timestamps besides the temporal distances. Under this circumstance, we propose REPLAY, learning to capture the time-varying temporal regularities for location prediction based on general RNN architecture. Specifically, REPLAY is designed on top of a flashback mechanism, where the spatiotemporal distances in sparse trajectories are used to search for the informative past hidden states; to accommodate the time-varying temporal regularities, REPLAY incorporates smoothed timestamp embeddings using Gaussian weighted averaging with timestamp-specific learnable bandwidths, which can flexibly adapt to the temporal regularities of different strengths across different timestamps. We conduct a comprehensive evaluation, comparing REPLAY against a wide range of state-of-the-art methods. Experimental results show REPLAY significantly and consistently outperforms state-of-the-art methods by 7.7\%-10.5\% in the location prediction task, and the learnt bandwidths reveal interesting patterns of the time-varying temporal regularities.
\end{abstract}

}

\maketitle

\IEEEdisplaynontitleabstractindextext

%
\IEEEpeerreviewmaketitle

\IEEEraisesectionheading{\section{Introduction}\label{sec:introduction}}

\IEEEPARstart Location prediction plays a crucial role in the modeling of human mobility, serving as a fundamental building block for various location based services. The primary aim is to forecast which location a user is most likely to visit given the user's historical data \cite{noulas2012mining}. As a sequence modeling problem by nature, location prediction problems have been widely tackled by the literature using Recurrent Neural Networks (RNNs), such as vanilla RNN \cite{mikolov2010recurrent}, Gated Recurrent Unit (GRU) \cite{cho2014learning} and Long Short-Term Memory (LSTM) \cite{hochreiter1997long}; these techniques have shown great success in modeling language data. However, different from word sequences (i.e., text sentences), real-world user mobility traces are often \textit{sparse and incomplete}, due to the nature of the data collection scheme and privacy issues. More precisely, Location Based Social Networks (LBSNs) are valuable data sources for benchmarking location prediction techniques, where users voluntarily share their spatiotemporal presence (check-ins) at specific Points of Interest (POIs) and specific times. However, due to the \emph{voluntary} sharing, the resulting mobility traces tend to be sparse in nature. For example, an empirical investigation using the commonly employed Foursquare dataset shows that the median time interval between consecutive check-ins is approximately 16.72 hours \cite{deng2023robust}. This sparsity and incompleteness in data present challenges for applying RNNs to the task of location prediction.

To tackle this sparsity issue, existing works strive to extend RNN architectures by incorporating spatiotemporal contexts, which have been shown as strong predictors for location prediction \cite{gonzalez2008understanding}. A widely adopted approach is to add the spatiotemporal distances between successive check-ins as additional inputs together with sequences of POIs to RNN units \cite{liu2016predicting,neil2016phased,zhu2017next,kong2018hst,cui2019distance2pre,zhao2019go,zhao2020go}. Despite its wide adoption, recent studies have shown that this approach cannot fully capture the temporal periodicity and spatial regularity of human mobility patterns \cite{yang2020location1}, as considering spatiotemporal distances between check-ins that are several steps apart (i.e., high-order distances) in the check-in sequences are often more informative than those between successive check-ins for location prediction. Subsequently, a flashback mechanism has been introduced to explicitly use such spatiotemporal distances in sparse trajectories search for the informative past hidden states (i.e., historical hidden states with similar spatiotemporal contexts to the current one) \cite{yang2020location1,li2021location,cao2021attention,rao2022graph,liu2022real,wu2022have,ye2023adaptive,yang2023uptdnet,ye2023adaptive}. However, despite the improved performance, the temporal distances still fail to capture the time-varying temporal regularities of human mobility, as we discuss below.

\begin{figure}
\centering
\includegraphics[width=\columnwidth]{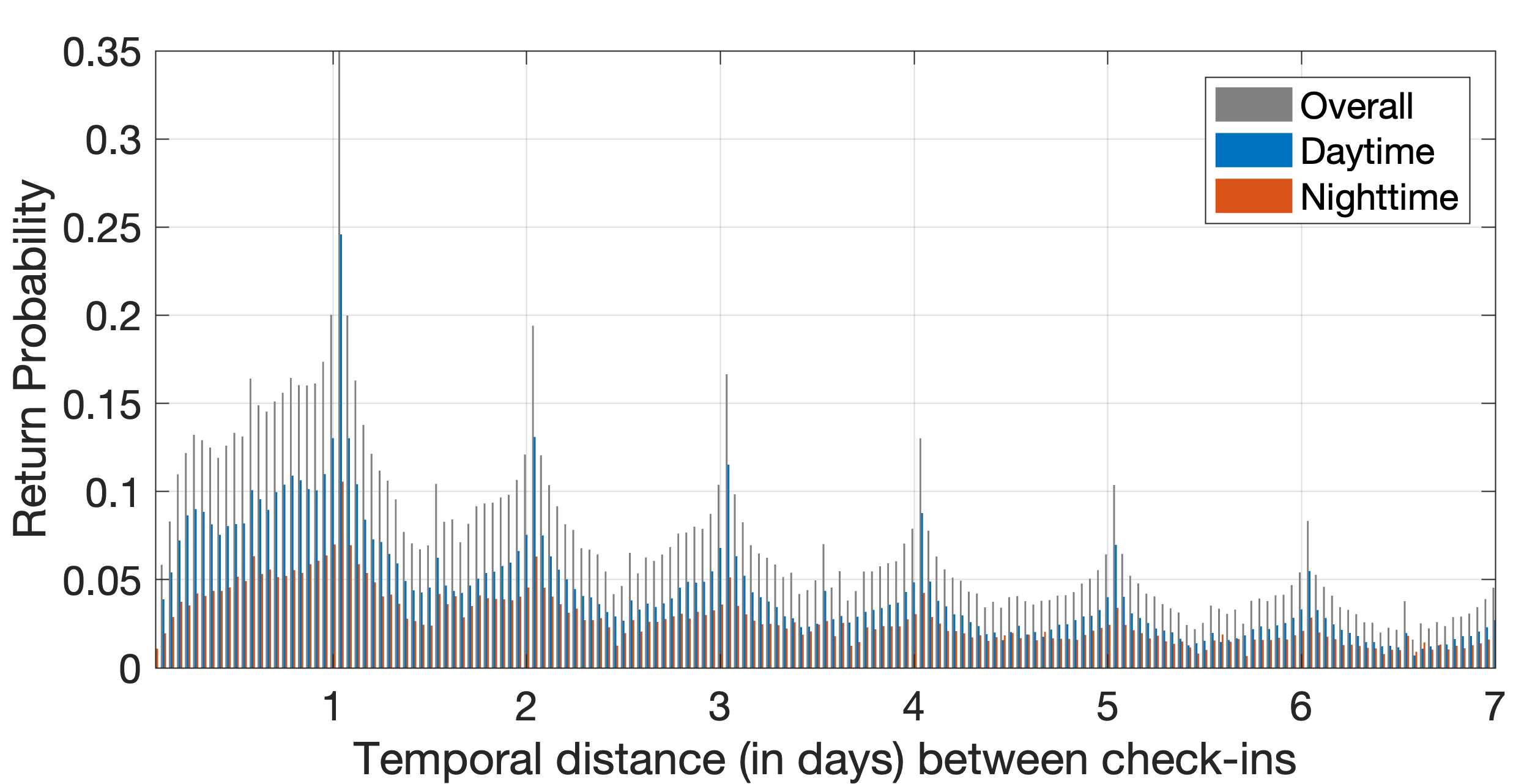}
\caption{Returning probability w.r.t. the temporal distances between check-ins. We plot the overall returning probability and its breakdown in the daytime (6:00-18:00) or nighttime (18:00-6:00). We observe that the daytime returning probability is significantly higher than in nighttime, which implies that the temporal regularity is much stronger in the daytime than in the nighttime.}
\label{example_time_regularity}
\end{figure}

Temporal regularity has been revealed as a universal law of human mobility \cite{gonzalez2008understanding,noulas2012tale}, which can be evidenced by the periodicity of human activities even in sparse human mobility traces \cite{cho2011friendship}. Figure \ref{example_time_regularity} shows the temporal returning probability of user check-ins on the Foursquare dataset. This probability represents the probability of users revisiting a particular POI within a specific time interval after their initial check-in at that POI \cite{gonzalez2008understanding,lian2015cepr}. We observe an obvious daily (periodic) revisiting pattern, which implies that historical check-ins closer to these daily peaks in temporal distance have higher predictive power; this serves as the foundation for the flashback mechanism \cite{yang2020location1}. In this paper, we further reveal that \textit{the strength of the temporal regularities varies across different time periods during a day (or across different days in a week, e.g., weekdays v.s. weekends)}. On the Foursquare dataset, the mobility pattern in the daytime is mostly attributed to working-related activities, which is usually more regular than the nighttime mobility pattern that is mostly attributed to entertainment-related activities\footnote{Note that user check-ins on LBSNs are mostly for social sharing purposes, where check-ins at entertainment-related POIs are much more than those at home-related POIs during the nighttime \cite{yang2020location}.}. Figure \ref{example_time_regularity} also compares the daytime and nighttime returning probabilities, where the daytime returning probability is significantly higher than nighttime, implying that the temporal regularity is much stronger in the daytime than in the nighttime. This implies that the hidden state 24 hours back is more informative for location prediction if the current time is daytime rather than nighttime. However, such time-varying temporal regularities cannot be captured by the distance-based methods (i.e., merely considering the temporal distances), and thus require involving the actual timestamps of check-ins.


Motivated by this observation, we propose REPLAY, a general RNN based architecture learning to capture the va\underline{R}ying t\underline{E}m\underline{P}oral regu\underline{LA}rit\underline{Y} of human mobility for location prediction. Specifically, REPLAY is designed to seamlessly incorporate smoothed timestamp embeddings with learnable bandwidths into the flashback mechanism. More precisely, for each check-in in a check-in sequence, we first transform its time to an hour-in-week timestamp $t_{n}$, where $1 \leq n \leq 168$ (one of the 168 hours in a week) and then generate a smoothed timestamp embedding using Gaussian weighted averaging with a timestamp-specific learnable bandwidth $\sigma_n$ (controlling the width of the Gaussian weighting function centered at the timestamp $n$). Afterward, we concatenated the smoothed timestamp embedding with the corresponding POI embedding, which is then fed into an RNN with the flashback mechanism, where the spatiotemporal distances in sparse trajectories are used to search for the informative past hidden states. Finally, we combine the hidden state output by the RNN, a user embedding, and a smoothed timestamp embedding generated from a query time to make predictions on future POIs. \textit{The timestamp-specific learnable bandwidths here can automatically learn to adapt to the temporal regularities of different strengths across different timestamps.} Intuitively, a smaller $\sigma_n$ leads to a peaky Gaussian, which implies a stronger regularity at timestamp $n$, as the smoothed timestamp embedding relies mostly on the embedding of $t_n$ itself and less on the information from its neighboring timestamps for prediction. This is also evidenced by our experiments later; the learnt bandwidths for the timestamps in the daytime are indeed 7.5\% smaller on average than those in the nighttime, which well corresponds to our observations in Figure \ref{example_time_regularity}. We summarize our contributions as follows:

\begin{itemize}
    \item We revisit the existing sequence models for location prediction, and identify their limitation in capturing the time-varying temporal regularities of human mobility.
    \item We propose REPLAY, a general RNN architecture capturing the time-varying temporal regularities via smoothed timestamp embeddings using Gaussian weighted averaging with timestamp-specific learnable bandwidths, which can flexibly adapt to the temporal regularities of different strengths across different timestamps.
    \item We conduct a thorough evaluation using two real-world LBSN datasets. Results show that our REPLAY outperforms a sizable collection of cutting-edge location prediction methods, with 7.7\%-10.5\% improvement over the best-performing baselines. Moreover, the learnt bandwidths reveal interesting temporal regularity patterns: 1) morning mobility shows a stronger regularity compared to other time periods, which is consistent on both weekdays and weekends; 2) weekend mobility is less regular than weekdays in general, while the weekend afternoon sometimes shows the weakest regularity.
\end{itemize}

\section{Related Work}
\subsection{Human Mobility Trajectory}
Ravenstein published work on the analysis of migration patterns using human mobility trajectories derived from demographic data \cite{ravenstein1885laws} is pioneering work in studying human mobility. Nowadays, the widespread use of smart devices equipped with sensors has significantly enhanced the accessibility of human trajectory data for comprehensive studies \cite{doulkeridis2021survey}. Human trajectory data can be divided into two different types based on the data collection method as follows.

Firstly, there are \textit{continuously sampled} trajectories, which encompass regularly recorded location sequences obtained from devices equipped by individuals. For instance, Microsoft Research Asia conducted the GeoLife experiment \cite{zheng2008understanding} which collects trajectories from 182 volunteers for more than three years by capturing GPS traces every five seconds or ten meters. Another notable example is that IDIAP Research Institute and Nokia Research Center conducted the Lausanne Data Collection Campaign \cite{laurila2012mobile}, where the mobility traces of 200 individuals were collected. Each individual carried a smartphone that periodically recorded various sensor readings (e.g., Bluetooth, WLAN) at intervals such as every 60 seconds.  While these human trajectory datasets provide detailed information about individual mobility patterns, their scale is often limited due to designed settings and individuals' concerns regarding privacy (e.g., unwillingness to install software that continuously monitors and collects data on personal devices).

Secondly, there are \textit{voluntarily shared} trajectories, which consist of self-reported location sequences, primarily provided by users on social media network platforms. For instance, a large number of individuals willingly share their check-ins on LBSNs, contributing to the accumulation of a substantial collection of human trajectories. However, people often choose to disclose a portion of their check-ins by reporting at specific POIs. This means that users may visit various POIs without recording, influenced by factors such as their preferences for certain places, simply forgetting to do so, or even privacy concerns. Consequently, despite LBSNs being commonly acknowledged as an important data source for conducting extensive research and data collection, the resulting mobility trajectories inherently exhibit \textit{sparsity} \cite{guo2018learning, shi2020urbanmotion, yang2019revisiting}. This sparsity needs to be taken into careful consideration when conducting data analysis tasks \cite{wang2016will}. In light of this context, the focus of this paper is to investigate the problem of location prediction over such sparsely populated trajectories.

\subsection{Location Prediction}

The spatiotemporal regularity of human mobility has been revealed with ample empirical evidence in the literature \cite{gonzalez2008understanding}. Universal collective mobility patterns have been discovered across different cities using statistical methods \cite{noulas2012tale,oliveira2016regularity}, such as periodicity \cite{cho2011friendship}, preferential return \cite{song2010modelling}, activity preference \cite{yang2015modeling}, and location specificity \cite{yang2020location1}, which serve as the foundation for human mobility modeling. Location prediction plays an important role in human mobility study, which aims to predict the next location of users that they are most likely to visit by leveraging their historical traces. Exiting works widely leverage those universal mobility patterns as prior knowledge in designing location prediction models.

Traditional methods mainly focus on various mobility features including activity preferences \cite{ye2013s,yang2015modeling}, historical visit counts \cite{noulas2012mining,gao2012exploring}, social information \cite{cho2011friendship,sadilek2012finding}, or features (embedding) learnt by graph embedding techniques \cite{xie2016learning,feng2017poi2vec,qian2019spatiotemporal,yang2019revisiting}. In addition, factorization and generative methods have been utilized to tackle the challenges of location recommendation and prediction tasks \cite{kurashima2013geo,yang2013sentiment,yin2013lcars,lian2014geomf,wang2015regularity,chan2012utilizing}. However, these techniques possess limited ability to capture the sequential patterns that are inherent in human mobility, which have been proven to be crucial predictors for accurate location prediction \cite{liu2016predicting}.

To model the sequential patterns of human mobility, (Hidden) Markov Chain based methods have been extensively employed for sequential prediction \cite{mathew2012predicting,cheng2013you,feng2015personalized}. The fundamental idea is to calculate the transition probability of a particular state according to the preceding one. Factorizing Personalized Markov Chains (FPMC) is a typical technique employed in this context \cite{rendle2010factorizing}, which factorizes a personalized transition matrix in user-specific latent factors. To address the location prediction problem, FPMC has been extended by integrating spatial constraints into its modeling framework \cite{cheng2013you,feng2015personalized}. However, one limitation of these location prediction techniques is their inability to effectively capture long-term dependencies in human mobility trajectories.

To capture the long-term dependencies, Recurrent Neural Networks (e.g., RNN, GRU and LSTM) have been widely used in location prediction tasks. To handle real-world mobility traces that are often sparse and incomplete, spatiotemporal contexts are often considered in the RNNs. A common approach widely adopted is to augment the Recurrent Neural Network (RNN) units with the inclusion of temporal and spatial distances between consecutive check-ins as extra information. For instance, Distance2Pre \cite{cui2019distance2pre} considers the spatial distance between consecutive check-ins as extra information; STRNN \cite{liu2016predicting} extends RNN with time and spatial-specific transition matrices for location prediction; HST-LSTM \cite{kong2018hst} extends  LSTMs gates to consider spatiotemporal distance; STGN \cite{zhao2019go} adds spatial and temporal gates to the conventional LSTM to capture sequential patterns of users; NeuNext \cite{zhao2020go} jointly learn the location context prediction and the next location prediction tasks. STAN \cite{luo2021stan} uses spatiotemporal attention networks to capture the dependencies between non-adjacent check-ins; GeoSAN \cite{lian2020geography} uses hierarchical gridding of GPS locations for spatial discretization and uses self-attention layers for matching. Despite the improved performance of these methods, we have shown in study \cite{yang2020location1} that only considering the spatiotemporal distances between consecutive check-ins cannot fully capture the two universal human mobility laws, i.e., spatial regularity and temporal periodicity, for location prediction.

In this context, the flashback mechanism proposed by \cite{yang2020location1} explicitly utilizes high-order spatiotemporal distance information to find informative past hidden states. This flashback mechanism has been widely utilized by subsequent studies. For instance, BiGRU \cite{cao2021attention} incorporates the flashback network with a bi-directional GRU; BSDA \cite{li2021location} incorporates the flashback mechanism with an extra RNN which models correlations of user appealed by POIs from bi-direction speculation; Bi-STAN \cite{wu2022have} utilizes flashback mechanism for missing value imputation task; RTPM \cite{liu2022real} addresses real-time location recommendation task by combining flashback mechanism with real-time user preference; Graph-Flashback \cite{rao2022graph} incorporates knowledge graph embeddings technique into the flashback framework. Note that apart from RNNs, other sequential methods have also been utilized for location prediction, such as geography-aware self-attention networks \cite{lian2020geography,liu2024novel}, graph-enhanced attention networks \cite{li2021discovering,wang2022graph,ghosh2024mobilytics,lv2021private}, spatiotemporal attention networks \cite{luo2021stan}, and hierarchical multi-task graph recurrent networks \cite{lim2022hierarchical}. However, all of these existing methods overlook the time-varying temporal regularities of user mobility (as evidenced in Figure \ref{example_time_regularity}), which motivates us to design REPLAY that can automatically adapt to the temporal regularities of different strengths over time.

Compared to Flashback \cite{yang2020location1}, the current work REPLAY makes the following differences. First, beyond the temporal periodicity, we further reveal the time-varying temporal regularities of human mobility in this work. Second, to capture the time-varying temporal regularities for location prediction, we design REPLAY on top of Flashback by seamlessly incorporating smoothed timestamp embeddings with learnable bandwidths into RNNs with the Flashback mechanism, where Flashback is now one of the three building blocks of REPLAY (more detail below). Third, the new experiments show superior performance of REPLAY against state-of-the-art techniques; in particular, compared to Flashback, REPLAY shows a significant improvement of 21.8\%-43.9\% in location prediction tasks.

\begin{figure*}
\centering
\includegraphics[width=1.85\columnwidth]{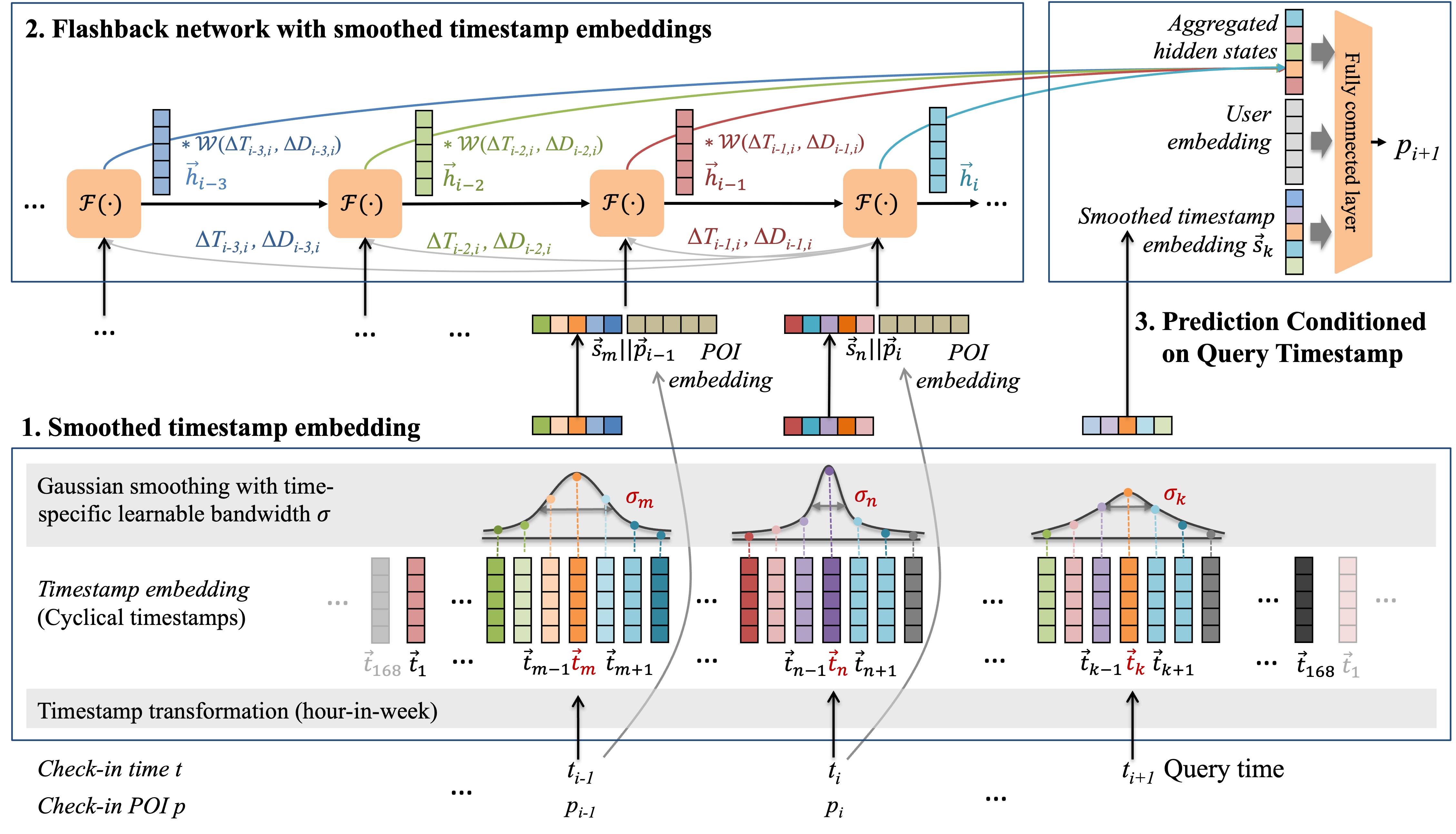}
\caption{Overview of our proposed REPLAY. It consists of three components: 1) Smoothed timestamp embedding, 2) Flashback network with smoothed timestamp embeddings, and 3) Prediction conditioned on query timestamp.}
\label{replay}
\end{figure*}

\section{REPLAY}
Motivated by the observation of the time-varying temporal regularities of human mobility, we design REPLAY to seamlessly incorporate smoothed timestamp embeddings with learnable bandwidths into the flashback mechanism. The timestamp-specific learnable bandwidths can automatically adapt to the temporal regularities of different strengths across different timestamps. Figure \ref{replay} shows the overall architecture of our proposed REPLAY. Given an individual's trace denoted as a sequence of POIs $\{..., p_{i-2}, p_{i-1}, p_{i},...\}$ and its associated check-in time $\{..., t_{i-2}, t_{i-1}, t_{i},...\}$, where $i$ denote the order of check-in in the mobility trace, our objective is to predict a future POI $p_{i+1}$ that an individual will probably visit at a query time $t_{i+1}$. To this end, REPLAY first generates a smoothed timestamp embedding for each check-in time in the sequence, and then concatenates it with the corresponding POI embedding which is then fed into an RNN with the flashback mechanism. Finally, we combine the hidden state output by the RNN, a user embedding, and a smoothed timestamp embedding generated from the query time to make predictions on future POIs conditioned on the query time. We present each component in detail below.

\subsection{Smoothed Timestamp Embeddings}
To flexibly capture the time-varying temporal regularities of human mobility, REPLAY designs a smoothed timestamp embedding method, which first transforms check-in time into hour-in-week timestamps, and then generates smoothed timestamp embeddings using Gaussian weighted averaging with learnable bandwidths, where cyclical timestamps are considered in the Gaussian kernel. 

\subsubsection{Hour-in-Week Timestamp Transformation}
For $i$-th check-in in the input mobility trace, we first transform its check-in time $t_{i}$ into an hour-in-week timestamp $t_{n}$, where $1 \leq n \leq 168$ (i.e., one of the 168 hours in a week). For example, $t_{i}=$ (15:00 Tue. 07/11/2023) is transformed into $t_{n}$, $n=39$, as $t_{i}$ is the 39-th hour in a week. We choose to model the temporal regularity at an hour granularity on a weekly scale, as it can capture not only daily patterns but also the difference between weekdays and weekends, which is also suggested by \cite{yang2019revisiting}. Moreover, we experimentally justify this design choice by comparing it with other time granularities and scales (see Section \ref{sec_exp_granu_scale} for more detail).

\subsubsection{Gaussian Smoothing with Learnable Bandwidth}
For each hour-in-week timestamp, we initialize a learnable embedding vector, denoted as $\vec{t}_{n}$, and then compute a smoothed timestamp embedding $\vec{s}_n$ using Gaussian weighted averaging with a learnable bandwidth $\sigma_n$, flexibly combining the neighboring timestamp embeddings. This design is inspired by kernel regression techniques \cite{nadaraya1964estimating}, and we further incorporate timestamp-specific learnable bandwidths here to flexibly capture the temporal regularities of different strengths across different timestamps. Specifically, for the input timestamp $t_n$, the Gaussian kernel function is defined as:
\begin{equation}
    f_n(l) = \frac{1}{\sigma_n \sqrt{2\pi}} e^{- \frac{dist(l,n)^2}{2\sigma_n^2}}
\end{equation}
where $dist(l,n)$ is a function computing the distance between timestamps $l$ and $n$. The bandwidth (i.e., standard deviation) $\sigma_n$ here controls the width of the Gaussian weighting function. A smaller bandwidth $\sigma_n$ leads to a peaky Gaussian, where the smoothed timestamp embedding of $t_n$ requires less information from its neighboring timestamps; in other words, the location prediction relies mostly on the timestamp embedding of $t_n$, which implies a stronger regularity at timestamp $n$. In contrast, a larger bandwidth $\sigma_n$ leads to a wide Gaussian, where the smoothed timestamp embedding of $t_n$ integrates a significant amount of information from its neighboring timestamps for location prediction, which thus implies a weaker regularity at timestamp $n$.

\subsubsection{Cyclical Timestamps}
Different from traditional distance metrics in kernel regression, our distance function for hour-in-week timestamps should be carefully designed, as Sunday night 11pm ($l=168$) and Monday morning 1am ($l=2$) are only two hours apart, for example. Therefore, we consider cyclical timestamps for the distance computation:
\begin{equation}
    dist(l,n) = \begin{cases}
|l-n|, &\text{if}\ |l-n|<84\\
168-|l-n|, &\text{otherwise}
\end{cases}
\end{equation}
Afterward, we compute the smoothed timestamp embedding $\vec{s}_n$ as a weighted average of all timestamp embeddings:
\begin{equation}
    \vec{s}_n = \frac{\sum_l f_n(l)\cdot \vec{t}_{l}}{\sum_l f_n(l)}
\end{equation}

\subsection{Flashback Network with Smoothed Timestamp Embeddings}
After obtaining the smoothed timestamp embedding, we feed it together with the corresponding POI embedding to RNN units with the flashback mechanism. In the following, we first briefly present the Flashback mechanism, followed by our extension of its input to add smoothed timestamp embeddings.

\subsubsection{Flashback Mechanism}

Flashback \cite{yang2020location1} is a general RNN architecture leveraging the spatiotemporal information in sparse trajectories to search the informative past hidden states for location prediction. Specifically, it takes a sequence of check-in POIs $\{..., p_{i-2}, p_{i-1}, p_{i},...\}$ as the input of an RNN (vanilla RNN, GRU, or LSTM) and output its recurrent hidden state $\vec{h}_i$. Then, the flashback mechanism leverages the spatiotemporal distance information to search the informative past hidden states. To this end, we use learnable weight $\mathcal{W}(\Delta T_{i,j}, \Delta D_{i,j})$ to re-weighting the historical hidden states $h_{j} (j<i)$ for location prediction. The weight used for aggregating the hidden state is parameterized by the spatial distance $\Delta D_{i,j}$ and temporal distance $\Delta T_{i,j}$ between the check-ins $(p_i, t_i)$ and $(p_j, t_j)$, which is defined as follows:

\begin{equation}
\mathcal{W}(\Delta T_{i,j}, \Delta D_{i,j}) = \mbox{hvc}(2\pi \Delta T_{i,j})  e^{-\alpha \Delta T_{i,j}} e^{-\beta \Delta D_{i,j}}
\label{equ_flashback}
\end{equation}
where the Havercosine function $\mbox{hvc}(x) = \frac{1+\cos(x)}{2}$ is used to capture the temporal periodicity ($\Delta T_{i,j}$ is measured in days here). This Havercosine function effectively models the daily periodicity of human mobility based on the temporal distances $\Delta T_{i,j}$ as shown in Figure \ref{example_time_regularity}. Note that the distance between transformed timestamps used in smoothed timestamp embeddings differs from the temporal distance in Flashback here; the former is used for smoothing timestamp embeddings to accommodate the time-varying temporal regularities, while the latter is used for weighing the past hidden state to capture the temporal periodicity. The two exponential terms model the temporal and spatial decay of past hidden states, respectively, following the intuition that the hidden states of older and farther check-ins have less predictive power in general; $\alpha$ and $\beta$ are tunable parameters for decay rates. Please refer to \cite{yang2020location1} for more details.

\subsubsection{Adding Smoothed Timestamp as Input}
Different from the original Flashback network \cite{yang2020location1} that only takes POI embeddings as inputs of RNN units, REPLAY concatenates the POI embedding together with the corresponding smoothed timestamp embedding as inputs of the RNN units, i.e., $\vec{h}_i = \mathcal{F}([\vec{p}_i;\vec{s}_n], \vec{h}_{i-1})$, as shown in Figure \ref{replay}. The objective here is to capture the correlation between a check-in's POI and timestamp. Such a correlation can effectively help the location prediction performance. For example, working-related POIs are usually checked by users in the daytime. Moreover, our smoothed timestamp embeddings here can \textit{flexibly} capture this correlation by integrating information from neighboring timestamps, where the amount of the information is controlled by the bandwidth $\sigma$. Subsequently, the learnable $\sigma$ can automatically adapt to time-varying temporal regularities; for example, a timestamp with a high mobility regularity will intuitively yield a smaller value of the bandwidth $\sigma$, as it relies less on the information from its neighboring timestamps for prediction.

\subsection{Prediction Conditioned on Query Timestamp}
To predict a future location, we feed the aggregated hidden state with a user embedding and a smoothed timestamp embedding of a query timestamp, to a fully connected layer to output a probability distribution of all POIs. 

The design choice of prediction conditioned on timestamp is motivated by the fact that over a sparse user mobility trace, the time intervals between successive check-ins significantly vary, which also implies that the next check-in will occur after an unknown period in the future. However, the predicted next location often varies depending on when in the future we would like to know the user’s location. For example, given a user's check-in sequence: visiting a bookstore at 10 AM, a restaurant at 11 AM, and a library at 3 PM, the predictions at 6 PM and 10 PM will probably differ (probably be a restaurant at 6 PM and be at home at 10 PM), as user behavior usually changes over time. Therefore, instead of making blind predictions under unknown prediction time, REPLAY is designed to make predictions conditioned on a ``query'' time (when we would like to know the user's location in the future). The same or similar settings exist in the literature by using a query timestamp (same as ours) \cite{gao2012exploring,yang2019revisiting,li2021location,feng2024rotan}, a query time interval (to predict the location of a user after a given time period) \cite{figueiredo2016tribeflow,zhao2019go,zhao2020go,feng2018deepmove}, or a fixed query time threshold (to predict the location of a user within a forthcoming period of time) \cite{feng2015personalized,zhang2014lore,feng2018deepmove}.




Note that an alternative setting is to make predictions on the next POI without considering the query time \cite{liu2016predicting,yang2020location1,rao2022graph}. We do not advocate for this setting in this paper as it always makes the same prediction for all future time given an input mobility trace, which is less reasonable in real-world scenarios.

\subsection{Model Training}
The training process of REPLAY follows the traditional backpropagation-through-time training process of recurrent neural networks \cite{werbos1990backpropagation}. For each prediction, we compute the cross-entropy loss between the probability distribution of POIs output by REPLAY and the ground truth POIs, which are minimized using Adam optimizer \cite{kingma2014adam}. The loss function is defined as follows:
\begin{equation}
    \mathcal{L} = -\frac{1}{U} \sum^{U}_{j=1} \frac{1}{N_j-1}  \sum^{N_j-1}_{i=1} y_{i+1,j} log(p(\hat{y}_{i+1,j}|\mathcal{H}_{i},s_{k},u_{j}))
\nonumber
\end{equation}
where $U$ is the number of users; $N_j$ is the sequence length of $j$-th user; $y_{i+1,j}$ is the one-hot vector of ground truth POI for ($i$+$1$)-th check-in of $j$-th user. $p(\hat{y}_{i+1,j}|\mathcal{H}_{i},s_{k},u_{j})$ is the predicted probability distribution given the current aggregated hidden state $\mathcal{H}_{i}$, smoothed query timestamp $s_{k}$ (transformed from time $t_{i+1}$), and user embedding $u_{j}$. For more training and implementation details of REPLAY, please refer to our code available online\footnote{https://github.com/UM-Data-Intelligence-Lab/REPLAY}.

\section{Discussions}
In this section, we discuss how our REPLAY addresses the sparsity issue of human trajectories through its key design ethos.

\begin{figure}
\centering
\includegraphics[width=\columnwidth]{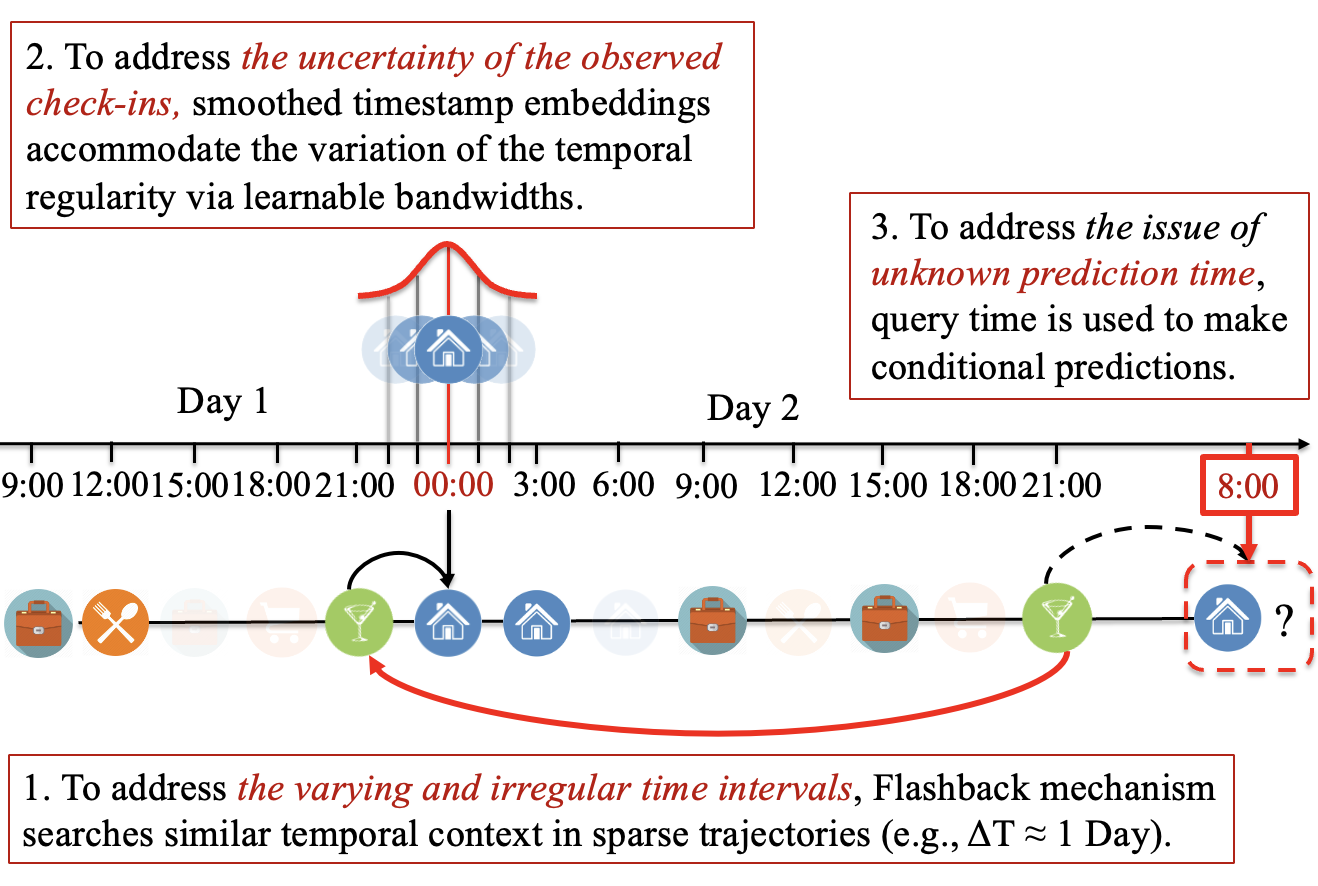}
\caption{A toy example showing the key design ethos of REPLAY addressing the sparsity of trajectories by tackling three key challenges 1) varying and irregular time intervals, 2) uncertainty of the observed check-ins, and 3) unknown prediction time, through the three key components 1) Flashback mechanism, 2) smoothed timestamp embeddings, and 3) the query time, respectively.}
\label{replay_example}
\end{figure}

First, the sparse trajectories imply irregularly observed check-ins, where the \textit{varying and irregular time intervals} between successive check-ins make the sequential patterns weak and thus hinder the application of traditional sequence models. To address this issue, REPLAY adopts the \textit{Flashback mechanism} which effectively utilizes the rich spatiotemporal context to search the historical hidden states for location prediction, where the historical hidden states temporally close to the periodicity peaks of returning probability (as shown in Figure \ref{example_time_regularity}) are usually more informative for location prediction. For example, Figure \ref{replay_example} shows that using time interval ($\Delta T \approx 1$ day), we can quickly find similar temporal context in sparse trajectories and retrieve the corresponding hidden state for prediction.

Second, the sparse trajectories imply the \textit{uncertainty of the observed check-ins}, where the regular activity of a user might be recorded at slightly different timestamps in a day in sparse trajectories, for example. We reveal in this paper that the degree of such uncertainty (or the strength of the temporal regularities) varies across time (e.g., across different time periods during a day as shown in Figure \ref{example_time_regularity}). To address this issue, REPLAY uses the \textit{smoothed timestamp embeddings} to accommodate such variations. Instead of directly using a specific timestamp embedding, we smooth it with the neighboring timestamps as shown in Figure \ref{replay_example}, where the timestamp-specific learnable bandwidths can automatically adapt to the temporal regularities of different strengths across different timestamps.

Third, the sparse trajectories imply the next check-in occurs after an unknown period in the future. Subsequently, traditional approaches often predict the next location under an \textit{unknown prediction time}, and thus always make the same prediction regardless of when the next check-in will occur, which is unrealistic in real-world scenarios. In contrast, the predicted next location should vary depending on when in the future we would like to know the user's location. To address this issue, REPLAY is designed to make predictions conditioned on a \textit{query time} as shown in Figure \ref{replay_example}, providing an additional clue for location prediction, which better fits real-world scenarios.



\section{Experiments}
We conduct an extensive evaluation of REPLAY on location prediction tasks. In the following, we start by presenting our experimental setup, followed by our results and discussions.

\subsection{Experiment Setup}
We present our experiment setup below, including datasets, baselines, evaluation metrics, and our experiment settings.

\subsubsection{Dataset}
We evaluate our model on two widely used real-world datasets: \textbf{Gowalla} \cite{cho2011friendship} and \textbf{Foursquare} \cite{yang2020location}, respectively. Table \ref{dataset} shows the dataset statistics. Each mobility trace is chronologically split into 80\% for training and 20\% for testing. 

Note that the original datasets use UTC time (Coordinated Universal Time) and it is inappropriate to directly use such time for modeling the temporal regularity of users worldwide, as the user mobility regularity depends on local time. We thus transform the UTC time of each check-in to a local time according to the time zone of the check-in location (GPS coordinates).

\begin{table}[]
\caption{Datasets Statistics}
\label{dataset}
\centering
\small
\begin{tabular}{lrr}
\hline
\textbf{Dataset}	&\textbf{Gowalla}		&\textbf{Foursquare} \\ \hline
\#Users	&52,979		&46,065 \\
\#POIs	&121,851		&69,005 \\
\#Checkins   &3,300,986		&9,450,342 \\
Collection period		&02/2009-10/2010	&04/2012-01/2014 \\ \hline
{\begin{tabular}[c]{@{}l@{}}Median time between \\successive check-ins \end{tabular}} &{\begin{tabular}[c]{@{}r@{}}11.09 hours \\ (0.46 days) \end{tabular}} 	&{\begin{tabular}[c]{@{}r@{}}16.72 hours \\ (0.70 days) \end{tabular}}	 \\ \hline
\end{tabular}
\end{table}

\subsubsection{Baselines}
Our baseline techniques can be divided into five types:
\begin{itemize}
\item \emph{User Preference-based Methods}. \textbf{WRMF} \cite{hu2008collaborative} converts implicit feedback into confidence values and employs alternating least squares for the computation of user and item latent factors.  \textbf{BPR} proposed by \cite{rendle2009bpr} is a recommendation approach specifically designed for implicit feedback based on ranking, which employs a pairwise ranking loss to learn user preferences; it exploits direct user-item interaction to separate negative items from positive items to alleviate the sparsity of data and improves the performance. For these two methods, We employ the default configurations of LibRec\footnote{https://github.com/guoguibing/librec}.


\item \emph{Feature-based methods}: Most Frequent Time (\textbf{MFT}) proposed by \cite{gao2012exploring} based on the observation that users often interact with POIs at specific times or during particular time intervals (24 hours a day) to rank a POI. \textbf{LBSN2Vec} \cite{yang2019revisiting} employs a random walk-with-stay scheme to automatically learn embeddings from an LBSN hypergraph; for location prediction tasks, LBSN2Vec ranks POIs by maximizing the cosine similarity with user and time embeddings.  In our experiments, the ratio of mobility data, the number of the window size of random walk, and the negative sample number are set to \{1, 10, 10\}, respectively.



\item \emph{Markov-Chain-based Methods}: \textbf{FPMC} \cite{rendle2010factorizing} uses matrix factorization techniques to factorize a personalized transition matrix in user-POI latent factors. The dimension of factorization and the negative sample numbers are set to \{32, 10\} in our experiments. \textbf{PRME} \cite{feng2015personalized} considers spatial constraints of human mobility and extends FPMC to location prediction problems; it captures the personalized POI mobility transition patterns by learning user-POI embeddings. The component weight, the hidden state size, and the threshold are set to \{0.2, 20, 360\} in our experiments. \textbf{TribeFlow} \cite{figueiredo2016tribeflow} predicts user trajectories using semi-Markov transition probabilities over latent space based on a mixture model. The number of transitions, cores,  batches, and iterations are set to \{0.3, 20, 20, 2000\}, respectively.

\item \emph{Basic RNNs}: \textbf{RNN} \cite{zhang2014sequential} is a basic Recurrent Neural Network architecture; for location prediction tasks, it captures sequential patterns from check-in datasets.
\textbf{LSTM} \cite{hochreiter1997long} is a variant of RNN, which contains three multiplicative gates and a memory cell to hold long-term dependencies. \textbf{GRU} \cite{cho2014learning} is equipped with two gates to control the information flow for the location prediction task. For all three RNN architectures and across all datasets, the hidden size is set to 10 in our experiments.



\item \emph{Spatiotemporal Sequence Models}: Variational Bayes RNNs (\textbf{VRNN}) \cite{chung2015recurrent}  use high-level latent random variables to model the variability and uncertainty in highly structured sequential data. For a fair comparison with our REPLAY, we further add the query timestamp to VRNN, denoted as \textbf{VRNN-QT}. The hidden size is set to 10 and the latent size is set to 5 in our experiments. \textbf{DeepMove} \cite{feng2018deepmove,feng2020predicting} proposes an attentional recurrent network for mobility prediction from lengthy and sparse trajectories. We employ default settings as suggested by the paper, and the embedding and the hidden size are set to \{50, 10\}, respectively. \textbf{STRNN} \cite{liu2016predicting} adopts time and spatial-specific transition matrices in RNN to model mobility patterns. \textbf{STGN} \cite{zhao2019go} adds spatial and temporal gates to the conventional LSTM to capture sequential patterns of users. \textbf{STGCN} \cite{zhao2019go} uses coupled input and forget gates in STGN. The hidden size is set to 10 in our experiment. \textbf {STAN} \cite{luo2021stan} utilizes self-attention layers along the trajectory to capture the point-to-point interaction between non-adjacent check-ins. The negative sample number, embedding size, and trajectory sequence are set to \{10, 50, 100\}, respectively. \textbf{GetNext} \cite{yang2022getnext} is a graph-enhanced transformer model that exploits user
global trajectory flow map for location prediction; for a fair comparison, the category of POIs are uniformly assigned to the same. The user embedding size, GCN layers' channel numbers, POIs embedding size, and transformer encoder embedding size are set to \{32, 64, 128, 1024\}, respectively. \textbf{Flashback} \cite{yang2020location1} uses a context-aware weighting mechanism to search informative past hidden states; \textbf{Graph-Flashback} \cite{rao2022graph} adds knowledge graph embeddings techniques to the flashback network, where the social relationships between users are also used. We set $\alpha$ and $\beta$ as suggested by Flashback in Graph-Flashback and Flashback.

\end{itemize}

\begin{table*}[]
\caption{Overall location prediction performance, where the best-performing ones are highlighted. $^\dagger$Methods make predictions using query time information as a query timestamp (MFT, LBSN2Vec, and REPLAY), as a query time interval (TribeFlow, DeepMove, STGN, STGCN, STAN), or as a query time threshold (PRME). Note that DeepMove predicts locations in a fixed future time slot, implicitly implying a fixed query time interval. The top two best-performing results are denoted in bold and underlined, respectively.}
\label{result}
\centering
\small
\begin{tabular}{l|l|rrrr|rrrr}
\hline
\multicolumn{2}{c|}{\multirow{2}{*}{ Method }} & \multicolumn{4}{c|}{Gowalla} & \multicolumn{4}{c}{Foursquare} \\\cline{3-10}

\multicolumn{2}{c|}{} & Acc@1 & Acc@5 & Acc@10 & MRR & Acc@1 & Acc@5 & Acc@10 & MRR \\
\hline
\multirow{2}{*}{ {\begin{tabular}[c]{@{}l@{}}User Preference \\based Methods \end{tabular}} } & WRMF & 0.0112 & 0.0260 & 0.0367 & 0.0178 & 0.0278 & 0.0619 & 0.0821 & 0.0427 \\
& BPR & 0.0131 & 0.0363 & 0.0539 & 0.0235 & 0.0315 & 0.0828 & 0.1143 & 0.0538 \\
\hline
\multirow{2}{*}{ {\begin{tabular}[c]{@{}l@{}}Feature-based \\ Methods \end{tabular}} } & MFT$^\dagger$ & 0.0525 & 0.0948 & 0.1052 & 0.0717 & 0.1945 & 0.2692 & 0.2788 & 0.2285 \\
& LBSN2Vec$^\dagger$ & 0.0864 & 0.1186 & 0.1390 & 0.1032 & 0.2190 & 0.3955 & 0.4621 & 0.2781 \\
\hline
\multirow{3}{*}{ {\begin{tabular}[c]{@{}l@{}}Markov-Chain \\ based Methods \end{tabular}} } & FPMC & 0.0479 & 0.1668 & 0.2411 & 0.1126 & 0.0753 & 0.2384 & 0.3348 & 0.1578 \\
& PRME$^\dagger$ & 0.0740 & 0.2146 & 0.2899 & 0.1504 & 0.0982 & 0.3167 & 0.4064 & 0.2040 \\
& TribeFlow$^\dagger$ & 0.0256 & 0.0723 & 0.1143 & 0.0583 & 0.0297 & 0.0832 & 0.1239 & 0.0645 \\
\hline
\multirow{3}{*}{ Basic RNNs } & RNN & 0.0881 & 0.2140 & 0.2717 & 0.1507 & 0.1824 & 0.4334 & 0.5237 & 0.2984 \\
& LSTM & 0.0621 & 0.1637 & 0.2182 & 0.1144 & 0.1144 & 0.2949 & 0.3761 & 0.2018 \\
& GRU & 0.0528 & 0.1416 & 0.1915 & 0.0993 & 0.0606 & 0.1797 & 0.2574 & 0.1245 \\
\hline
\multirow{8}{*}{ {\begin{tabular}[c]{@{}l@{}}Spatiotemporal \\ Sequence \\ Models \end{tabular}} } & VRNN  & 0.1029 & 0.2594 & 0.3296 & 0.1781 & 0.2105 & 0.4794 & 0.5677 & 0.3332 \\
& VRNN-QT$^\dagger$ & 0.1373 & 0.3054 & 0.3799 & 0.2180 & \underline{0.2835} & 0.5691 & 0.6429 & 0.4131 \\
& STRNN & 0.0900 & 0.2120 & 0.2730 & 0.1508 & 0.2290 & 0.4310 & 0.5050 & 0.3248 \\
& DeepMove$^\dagger$ & 0.0625 & 0.1304 & 0.1594 & 0.0982 & 0.2400 & 0.4319 & 0.4742 & 0.3270 \\
& STGN$^\dagger$ & 0.0624 & 0.1586 & 0.2104 & 0.1125 & 0.2094 & 0.4734 & 0.5470 & 0.3283 \\
& STGCN$^\dagger$ & 0.0546 & 0.1440 & 0.1932 & 0.1017 & 0.1878 & 0.4502 & 0.5329 & 0.3062 \\
& STAN$^\dagger$ & 0.0891 & 0.2096 & 0.2763 & 0.1523 & 0.2265 & 0.4515 & 0.5310 & 0.3420 \\
& GETNext & 0.0912 & 0.2003 & 0.2487 & 0.1484 & 0.1862 & 0.4702 &0.5763 & 0.3153 \\
& Flashback & 0.1158 & 0.2754 & 0.3479 & 0.1925 & 0.2496 & 0.5399 & 0.6236 & 0.3805 \\
& Graph-Flashback & \underline{0.1512} & \underline{0.3425} & \textbf{0.4256} & \underline{0.2422} & 0.2805 & \underline{0.5757} & \underline{0.6514} & \underline{0.4136} \\
& REPLAY$^\dagger$ & \textbf{0.1866} & \textbf{0.3476} & \underline{0.4124} & \textbf{0.2635} & \textbf{0.3529} & \textbf{0.5953} & \textbf{0.6648} & \textbf{0.4638} \\
\hline
\end{tabular}
\end{table*}

\subsubsection{Evaluation Metrics}
We adopt two widely used evaluation metrics: Mean Reciprocal Rank (MRR) and average Accuracy@N (Acc@N) with N = \{1, 5, 10\} in our experiments. MRR measures the rank of the correctly predicted POI in the ordered result list, while Acc@N indicates whether the true POI appears within the top-N predicted POIs. Their definitions are as follows:

\begin{equation}
\text{MRR} = \frac{1}{m} \sum_{i=1}^{m} \frac{1}{rank}
\end{equation}
\begin{equation}
\text{Acc@N} = \frac{1}{m} \sum_{i=1}^{m} \mathbbm{1} (rank \leq N)
\end{equation}
where $m$ is the number of predictions and $\mathbbm{1}$ is the indicator function that returns 1 if the condition is true, otherwise 0. The $rank$ represents the rank of the true POI in the predicted ordered list. For both metrics, the larger the value, the better the performance.

\subsubsection{Experiment Settings}
We developed our model using the PyTorch framework and conducted experiments on the following hardware platform (CPU: Intel(R) Xeon(R) Gold 5320, GPU: NVIDIA GeForce RTX 3090). We evaluate REPLAY and the baselines in the location prediction task, aiming to predict the next POI a user will visit, based on her historical check-in sequence and a query time, as illustrated in Figure \ref{replay}. We empirically set the dimension of hidden states and all (POI, timestamp, and user) embedding sizes as 10, and the temporal decay factor $\alpha$ and spatial decay factor $\beta$ following the suggested settings in \cite{yang2020location1}, i.e., $\alpha=0.1$ on both datasets, $\beta = 1000$ and $100$ on Gowalla and Foursquare datasets, respectively, and the distance to flashback as 20. We implement our REPLAY with different RNN architectures (including vanilla RNN, GRU and LSTM), and systematically compare their performance in an ablation study below; in other experiments, we report the results of REPLAY using the best RNN architecture setting (vanilla RNN), if not specified otherwise.



\subsection{Location Prediction Performance}
Table \ref{result} shows overall location prediction performance. First, we find that REPLAY achieves the best performance in most cases, yielding 7.7\% and 10.5\% improvement over the best-performing baselines on Gowalla and Foursquare, respectively. Moreover, REPLAY significantly outperforms the best baselines in Acc@1 (by 23.4\% and 25.8\% on Gowalla and Foursquare, respectively). The only exception is on Acc@10 where Graph-Flashback achieves slightly better performance than REPLAY; note that beyond user mobility trajectories, Graph-Flashback further uses user social relationships as input.

We also highlight in Table \ref{result} the methods that make predictions using query time information in different formats (e.g., a query timestamp, a query time interval, or a query time threshold). However, we did not observe a consistent improvement of these methods over others that make predictions without using query time information. This is due to the fact that the distinct models adopted by these baseline methods dominate the prediction performance. Moreover, the query time information is often used by existing methods on an ad-hoc basis without a systematic study of its impact. Notably, compared to VRNN-QT (with the same query time as our REPLAY) but using the Variational Bayes component in RNNs to model the uncertainty of check-in time, our REPLAY shows 19.8\% and 11.2\% improvement on Gowalla and Foursquare,
respectively. The superiority of REPLAY over VRNN-QT implies our Gaussian weighted averaging method better captures the irregularity and uncertainty of check-ins than Variational Bayes RNNs.

In the following, we conduct a systematic ablation study below to verify the key design choices of our REPLAY.

\begin{table*}[]
\caption{Ablation study on REPLAY}
\label{ablation_study}
\centering
\small
\begin{tabular}{l|cccc|cccc} 
\hline
\multirow{2}{*}{Method} & \multicolumn{4}{c|}{Gowalla} & \multicolumn{4}{c}{Foursquare} \\ \cline{2-9}
& Acc@1 & Acc@5 & Acc@10 & MRR & Acc@1 & Acc@5 & Acc@10 & MRR \\ \hline
Flashback (RNN) & 0.1158 & 0.2754 & 0.3479 & 0.1925 & 0.2496 & 0.5399 & 0.6236 & 0.3805 \\
Flashback (LSTM) & 0.1024 & 0.2575 & 0.3317 & 0.1778 & 0.2398 & 0.5169 & 0.6014 & 0.3654 \\
Flashback (GRU) & 0.0979 & 0.2526 & 0.3267 & 0.1731 & 0.2375 & 0.5154 & 0.6003 & 0.3631 \\ \hline
REPLAY-noSTE (RNN) & 0.1362 & 0.3099 & 0.3845 & 0.2185 & 0.3055 & 0.5690 & 0.6415 & 0.4251 \\
REPLAY-noSTE (LSTM) & 0.1263 & 0.2980 & 0.3724 & 0.2079 & 0.3002 & 0.5663 & 0.6396 & 0.4209 \\
REPLAY-noSTE (GRU) & 0.1356 & 0.3099 & 0.3856 & 0.2183 & 0.2993 & 0.5634 & 0.6360 & 0.4192 \\ \hline
REPLAY-noQT (RNN) & 0.1397 & 0.3169 & 0.3924 & 0.2237 & 0.2817 & 0.5744 & 0.6489 & 0.4139 \\
REPLAY-noQT (LSTM) & 0.1343 & 0.3123 & 0.3873 & 0.2187 & 0.2739 & 0.5620 & 0.6360 & 0.4043 \\
REPLAY-noQT (GRU) & 0.1319 & 0.3105 & 0.3866 & 0.2168 & 0.2716 & 0.5623 & 0.6376 & 0.4029 \\ \hline
REPLAY-MultiG (RNN) & 0.1856 & 0.3357 & 0.3942 & 0.2570 & 0.3501 & 0.5894 & 0.6578 & 0.4598 \\
REPLAY-MultiG (LSTM) & 0.1808 & 0.3314 & 0.3904 & 0.2527 & 0.3440 & 0.5816 & 0.6493 & 0.4529 \\
REPLAY-MultiG (GRU) & 0.1798 & 0.3310 & 0.3897 & 0.2520 & 0.3426 & 0.5797 & 0.6477 & 0.4513 \\
\hline
REPLAY-FixedB(RNN) & 0.1861 & 0.3419 & 0.4024 & 0.2601 & 0.3513 & 0.5899 & 0.6575 & 0.4607 \\
REPLAY-FixedB (LSTM) & 0.1837 & 0.3411 & 0.4032 & 0.2587 & 0.3438 & 0.5817 & 0.6498 & 0.4528 \\
REPLAY-FixedB (GRU) & 0.1817 & 0.3374 & 0.3991 & 0.2558 & 0.3418 & 0.5807 & 0.6487 & 0.4512 \\
\hline
REPLAY (RNN) & \textbf{0.1866} & \textbf{0.3476} & \textbf{0.4124} & \textbf{0.2635} & \textbf{0.3529} & \textbf{0.5953} & \textbf{0.6648} & \textbf{0.4638} \\
REPLAY (LSTM) & 0.1848 & 0.3467 & 0.4100 & 0.2618 & 0.3448 & 0.5850 & 0.6544 & 0.4549 \\
REPLAY (GRU) & 0.1826 & 0.3449 & 0.4091 & 0.2597 & 0.3430 & 0.5831 & 0.6524 & 0.4530 \\
\hline
\end{tabular}
\end{table*}

\subsection{Ablation Study}
\label{sec_ablation}
We consider the following variation of REPLAY and results are shown in Table \ref{ablation_study}.
\begin{itemize}
    \item \textbf{REPLAY-noSTE} is a variant of REPLAY without the Smoothed Timestamp Embeddings (noSTE). It is also equivalent to Flashback by adding the query time interval for prediction. 
    \item \textbf{REPLAY-noQT} is a variant of REPLAY making predictions without using the Query Time (noQT). It is also equivalent to Flashback integrated with the smoothed timestamp embeddings.
    \item \textbf{REPLAY-MultiG} is a variant of REPLAY combining multi-granularity (hour-in-day and day-in-week) timestamp embeddings. 
    \item \textbf{REPLAY-FixedB} is a variant of REPLAY with a universal fixed non-learnable bandwidth.
    \item \textbf{Flaskback} can also be considered as a variant of REPLAY without the smoothed timestamp embeddings or query time.

\end{itemize}

For each method, we instantiate it with the three common RNN architectures (vanilla RNN, LSTM, or GRU). First, we notice that REPLAY outperforms REPLAY-noSTE by 20.6\% and 7.5\% on Gowalla and Foursquare, respectively. This demonstrates our smoothed timestamp embeddings indeed boost performance significantly by flexibly capturing the time-varying temporal regularities of human mobility. Second, we find that REPLAY outperforms REPLAY-noQT by 18.0\% and 11.1\% on Gowalla and Foursquare, respectively,  which shows the effectiveness of our prediction conditioned on query time as the query time intuitively provides additional information for location prediction problems. Moreover, REPLAY significantly outperforms Flashback by 43.9\% and 21.8\% on Gowalla and Foursquare, respectively, which verifies the effectiveness of integration smoothed timestamp embeddings and making predictions conditioned on query time.

To future explore the impact of different granularities and the timestamp-specific learnable bandwidths, we conducted experiments on two variants of REPLAY, named REPLAY-MultiG (\underline{Multi}-\underline{G}ranularity) and REPLAY-FixedB (\underline{Fixed} \underline{B}andwidth). From Table \ref{ablation_study}, we observe that REPLAY consistently yields better performance (3.4\% and 0.6\% improvement on Gowalla and Foursquare, respectively) over REPLAY-MultiG, which further validates our design choice of hour-in-week timestamps. Besides, we find that REPLAY with timestamp-specific learnable bandwidths consistently outperforms, with improvements of 1.5\% on Gowalla and 0.6\% on Foursquare compared to REPLAY-FixedB. This highlights the importance of utilizing timestamp-specific learnable bandwidths in location prediction.

In addition, we also investigate the performance of using different RNN architectures (vanilla RNN, GRU or LSTM). We observe that the vanilla RNN yields the best performance in general, while LSTM and GRU show comparable results. This is probably due to the fact that the flashback mechanism is able to effectively find useful historical hidden states using spatiotemporal contexts for location prediction, which indeed weakens the utility of retaining long-term memory by LSTM or GRU. 

Interestingly, we find that REPLAY is more robust than Flashback against different RNNs. Specifically, we use the coefficient of variation \cite{everitt2010cambridge} to evaluate robustness, which is the standard deviation ratio to a variable's mean. We compute the coefficient of variation on the performance using different RNNs, and compare the results from different methods. Table \ref{res_cov} shows that compared to basic RNNs, Flashback can effectively reduce the coefficient of variation from 21.73\% and 41.84\% to 5.59\% and 2.55\% on Gowalla and Foursquare, respectively. More importantly, REPLAY can further reduce the coefficient of variation to 0.73\% and 1.26\% on Gowalla and Foursquare, respectively. In other words, integrating smoothed timestamp embeddings and making predictions conditioned on query time can make our model more robust against different RNNs.

\begin{table}[]
\caption{Coefficient of variation in MRR over Different RNN Architectures
(RNN, GRU, LSTM).}
\label{res_cov}
\centering
\small
\begin{tabular}{l|r|r} \hline
Method & Gowalla & Foursquare \\ \hline
Basic RNNs & 21.73\% & 41.84\% \\
Flashback & 5.59\% & 2.55\% \\
REPLAY-noSTE & 2.82\% & 0.71\% \\
REPLAY-noQT & 1.64\% & 1.47\% \\
REPLAY-MultiG  &1.06\% & 0.99\%  \\ 
REPLAY-FixedB   &0.84\%  &1.12\%  \\
REPLAY & 0.73\% & 1.26\% \\ \hline
\end{tabular}
\end{table}



\begin{table}[]
\caption{Settings on Time Granularities and Scales}
\label{time_granularity_setting}
\centering
\small
\begin{tabular}{l|l|r} \hline
Scale & Granularity & \#Timestamps \\ \hline
\multirow{2}{*}{ Day } & Minute & 1,440 \\ \cline{2-3}
& Hour & 24 \\ \hline
\multirow{2}{*}{ {\begin{tabular}[l]{@{}l@{}}Weekday\& \\ Weekend \end{tabular}} } & Minute & 2,880 \\ \cline{2-3}
& Hour & 48 \\ \hline
\multirow{2}{*}{ Week } & Minute & 10,080 \\ \cline{2-3}
& Hour & 168 \\ \hline
\end{tabular}
\end{table}

\begin{table*}[]
\caption{Impact of different time granularities and scales on location prediction performance}
\label{time_granularity_res}
\centering
\small
\setlength{\tabcolsep}{0.6em}
\begin{tabular}{l|l|cccc|c|cccc|c}
\hline
\multirow{3}{*}{Scale} & \multirow{3}{*}{Granularity} & \multicolumn{5}{c|}{Gowalla} & \multicolumn{5}{c}{Foursquare} \\ \cline{3-12}
& & Acc@1 & Acc@5 & Acc@10 & MRR & {\begin{tabular}[c]{@{}c@{}}Time per\\epoch \end{tabular}} & Acc@1 & Acc@5 & Acc@10 & MRR & {\begin{tabular}[c]{@{}c@{}}Time per\\epoch \end{tabular}} \\
\hline
\multirow{2}{*}{Day } & Minute & 0.1278 & 0.2918 & 0.3610 & 0.2060 & 41.35s & 0.2801 & 0.5723 & 0.6468 & 0.4121 & 122.14s \\
\cline{2-12}
& Hour & 0.1777 & 0.3411 & 0.4063 & 0.2555 & 39.35s & 0.3483 & 0.5927 & 0.6622 & 0.4601 & 101.85s \\
\hline
\multirow{2}{*}{{\begin{tabular}[l]{@{}l@{}}Weekday\& \\ Weekend \end{tabular}} } & Minute & 0.1804 & 0.3423 & 0.4090 & 0.2577 & 44.98s & 0.3530 & 0.5938 & 0.6631 & 0.4633 & 142.74s \\
\cline{2-12}
& Hour & 0.1821 & 0.3442 & 0.4097 & 0.2594 & 37.69s & 0.3505 & 0.5941 & 0.6632 & 0.4618 & 102.33s \\
\hline
\multirow{2}{*}{Week } & Minute & 0.1756 & 0.3276 & 0.3891 & 0.2485 & 72.41s & \textbf{0.3532} & 0.5940 & 0.6638 & 0.4636 & 264.00s \\
\cline{2-12}
& Hour & \textbf{0.1866} & \textbf{0.3476} & \textbf{0.4124} & \textbf{0.2635} & 44.20s & 0.3529 & \textbf{0.5953} & \textbf{0.6648} & \textbf{0.4638} & 103.49s \\
\hline
\end{tabular}
\end{table*}

\subsection{Temporal Regularities on Different Time Granularities and Scales}
\label{sec_exp_granu_scale}
We investigate the impact of different time granularities (i.e., minute or hour) and time scales (i.e., day, weekday\&weekend or week) in our smoothed timestamp embeddings used by REPLAY. Table \ref{time_granularity_setting} shows the settings we considered and their corresponding number of timestamps. Note that for the time scale weekday\&weekend, we define two sets of daily cyclical timestamps for weekday and weekend, respectively; the similarity between the timestamps in each set is computed independently from the other set. This setting is motivated by the fact that human mobility patterns are often split into weekday and weekend patterns by existing studies \cite{oliveira2016regularity}. The performance comparison between different settings is shown in Table \ref{time_granularity_res}. 

First, we observe that the week scale is consistently better than the day scale. For example, on the hour granularity (the best time granularity setting as we discuss below), the week scale outperforms the day scale by 2.89\% and 0.74\% on Gowalla and Foursquare, respectively. This is due to the fact that the former can capture the different temporal regularities across different days in a week, in particular the different regularities on weekdays and weekends (more detail in Section \ref{sec_exp_temp_reg} below), while the latter cannot. Moreover, compared to the weekday\&weekend scale which can also capture the different regularities on weekdays and weekends, the week scale still achieves consistently better performance (with 1.43\% and 0.4\% improvement on Gowalla and Foursquare, respectively, under the hour granularity). Because compared to the weekday\&weekend scale, the week scale can subtly capture the regularity differences between different weekdays (also between different days in the weekend).

Second, we observe that the hour granularity is generally better than the minute granularity, as the minute granularity is often over-specific to model the temporal regularity of user mobility. One exception is on the week scale on Foursquare where we observe comparable results; however, in this case, the minute granularity shows significantly worse runtime performance (more than twice the training time as shown in Table \ref{time_granularity_res}) than the hour granularity, as it needs to learn much more embeddings for minute-in-week timestamps (10,080 in total) than for hour-in-week timestamps (168 in total). Moreover, we also conducted experiments on multi-granularity combining both hour-in-day and day-in-week timestamp embeddings in Section \ref{sec_ablation}; the results show 3.4\% and 0.6\% improvement of our REPLAY over this multi-granularity setting, on Gowalla and Foursquare, respectively. In summary, we choose hour-in-week timestamps in REPLAY as the default setting.


\begin{figure*}
\centering
\subfigure[Hours in a week] { 
\includegraphics[width=\columnwidth]{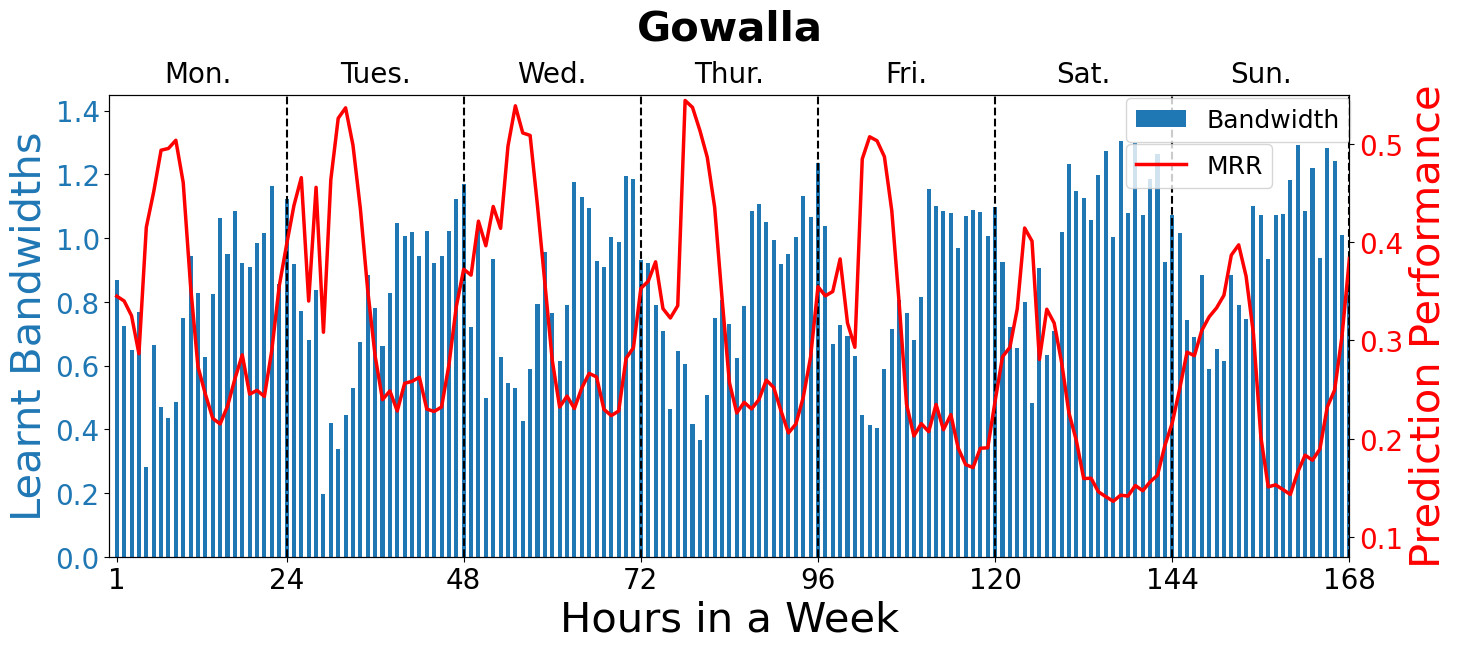}
\includegraphics[width=\columnwidth]{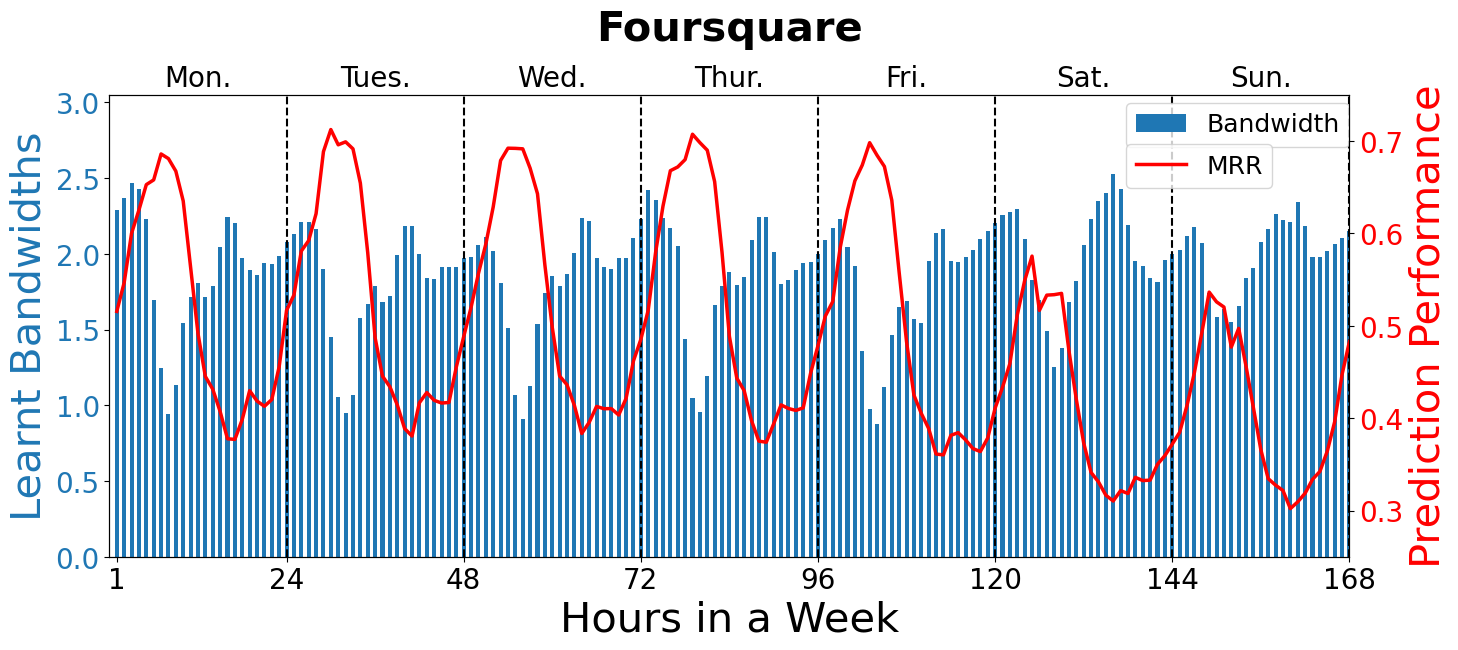}
\label{week_sigma}
}
\subfigure[Hours on weekdays] {
\includegraphics[width=0.48\columnwidth]{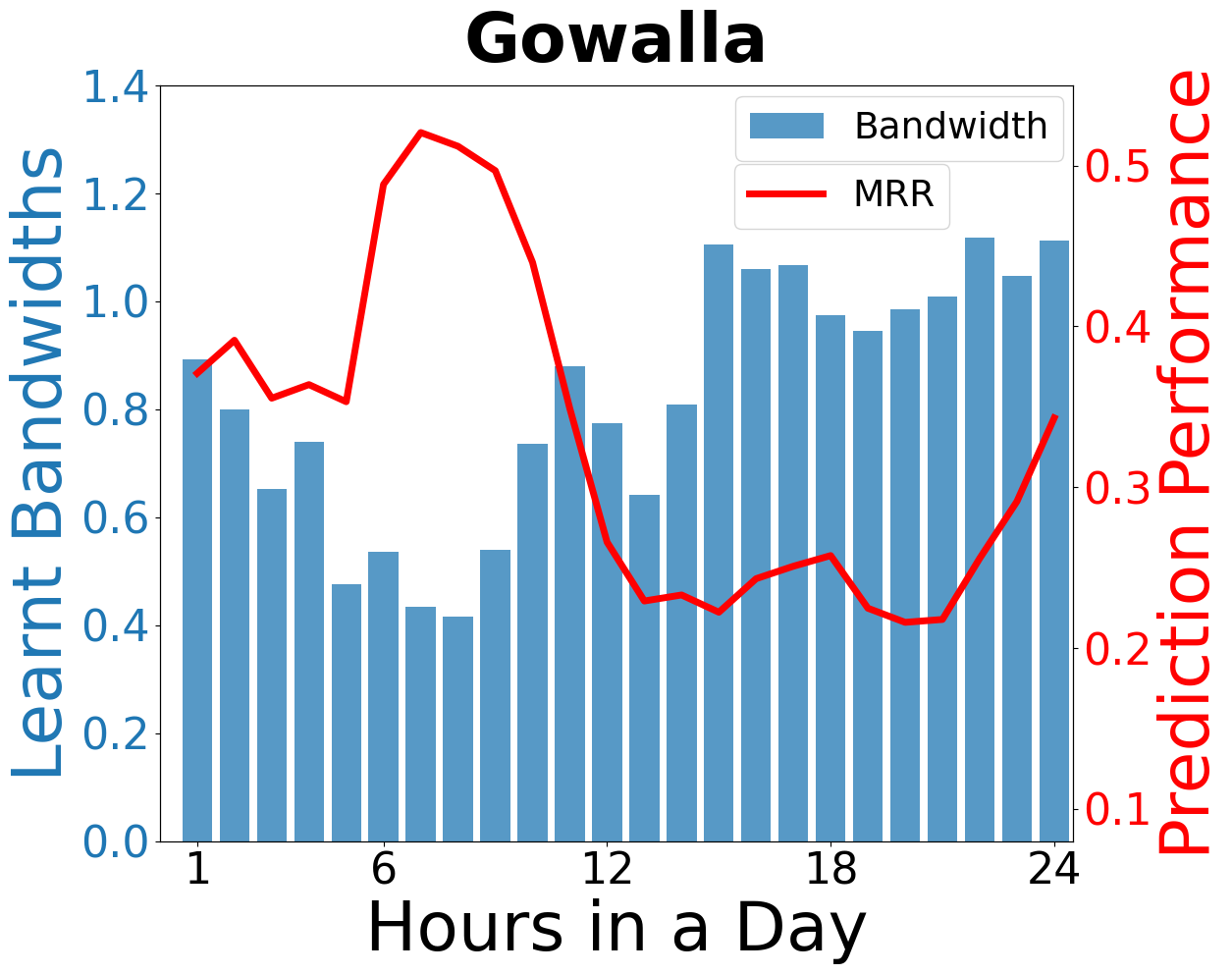}
\includegraphics[width=0.48\columnwidth]{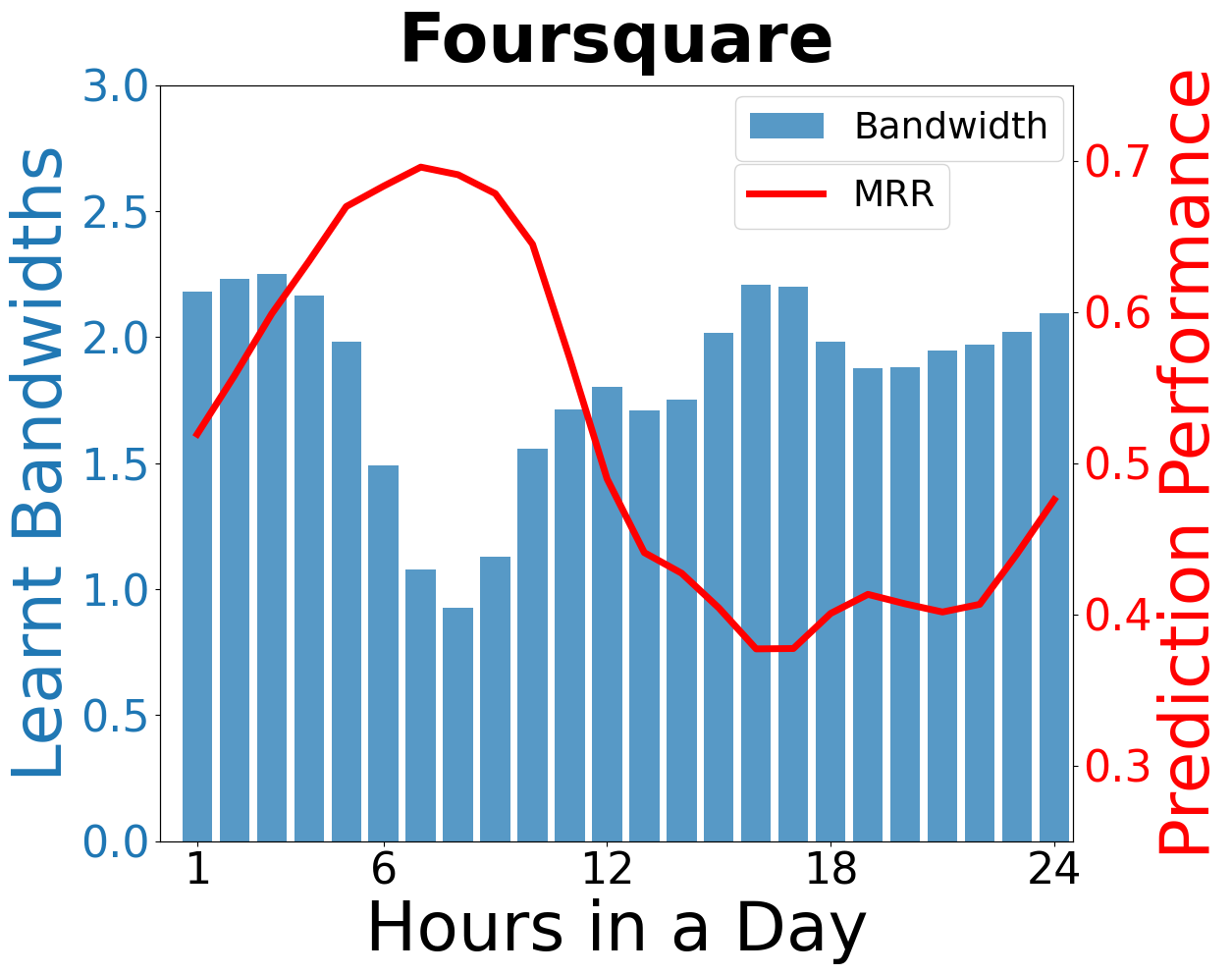}
\label{weekday_sigma}
}
\subfigure[Hours on weekends] {
\includegraphics[width=0.48\columnwidth]{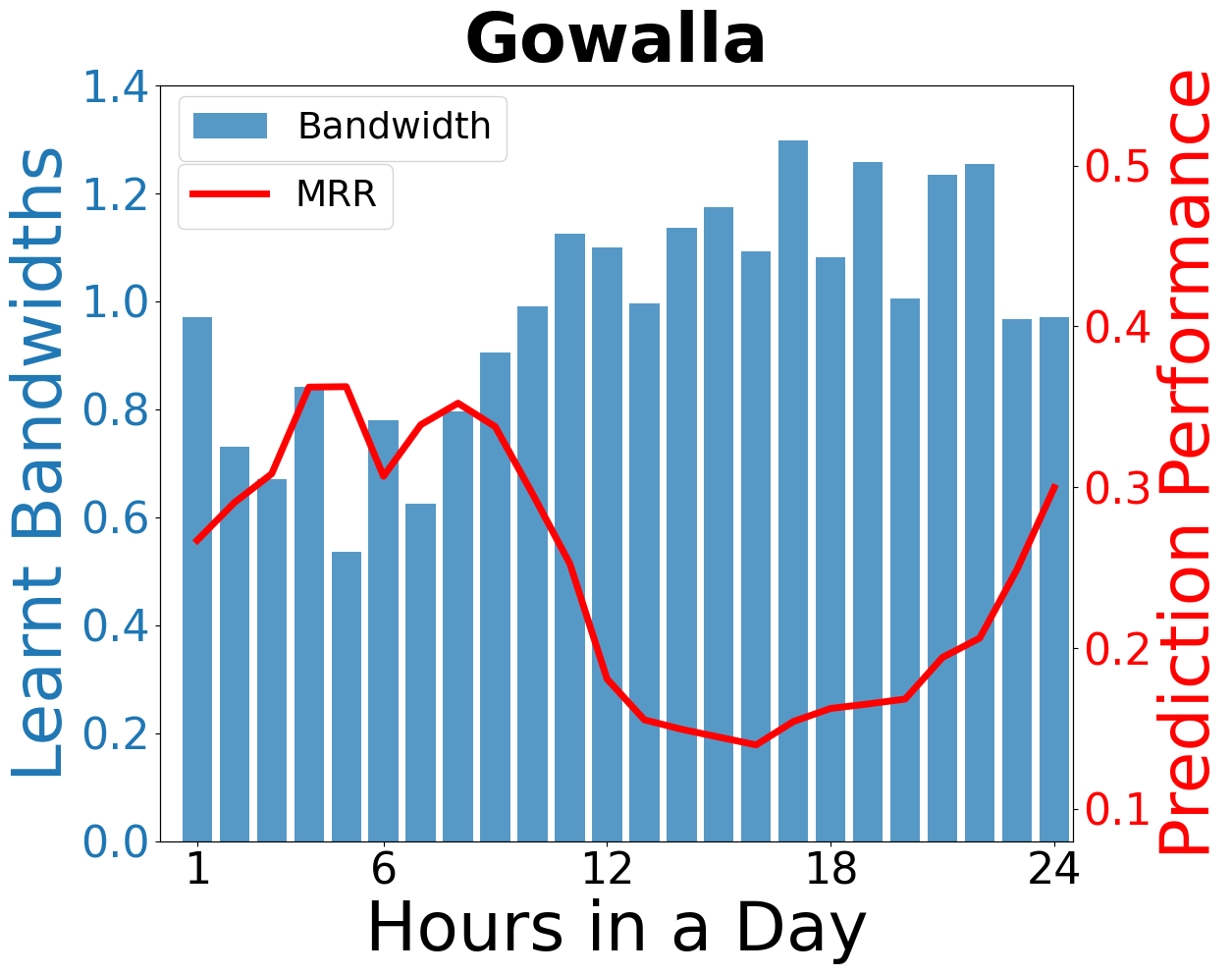}
\includegraphics[width=0.48\columnwidth]{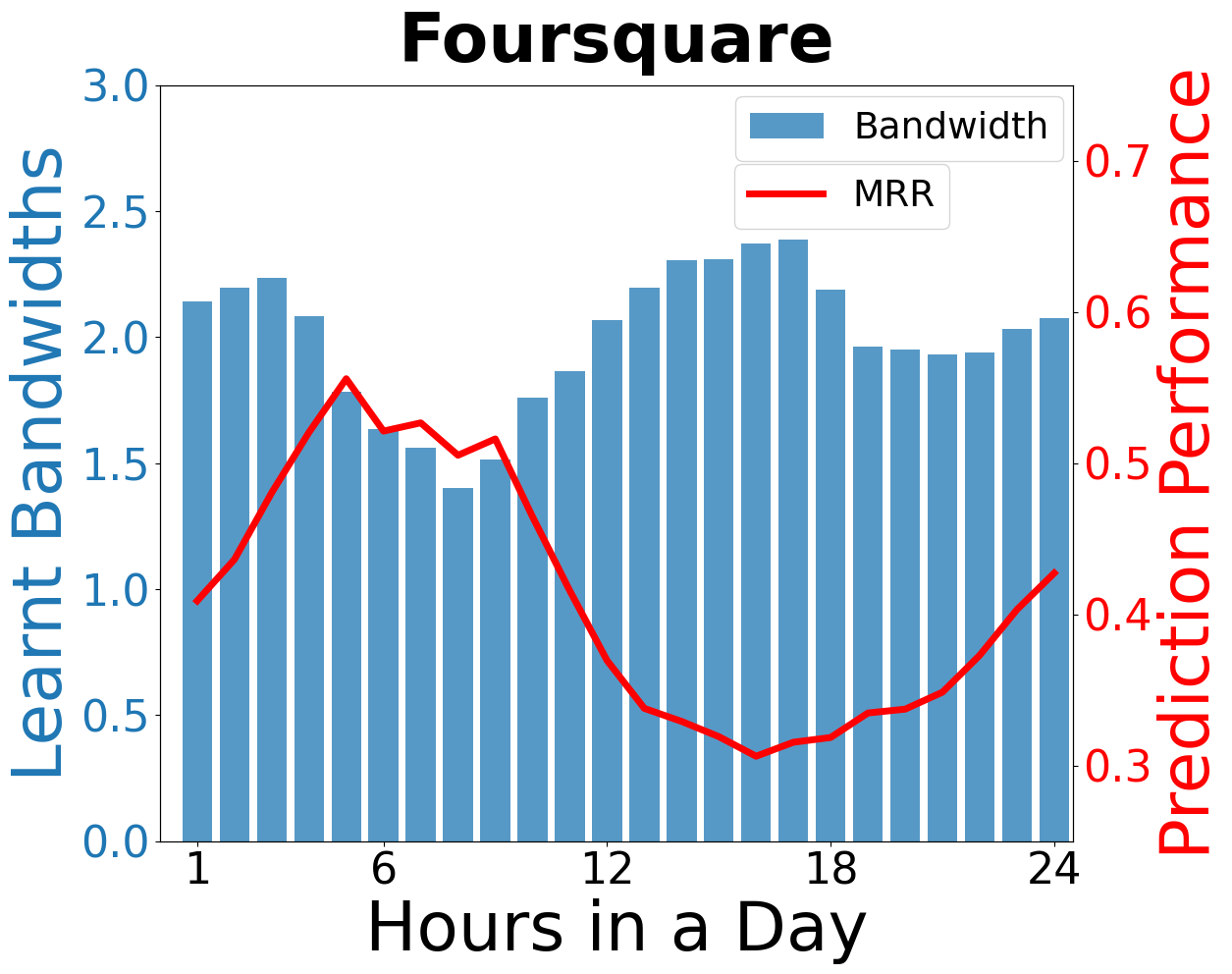}
\label{weekend_sigma}
}
\caption{The learnt bandwidths and the corresponding location prediction performance across different timestamps. }
\label{all_sigma}
\end{figure*}

\subsection{Revealing the Time-Varying Temporal Regularities}
\label{sec_exp_temp_reg}
In this experiment, we investigate the time-vary temporal regularities of user mobility using the learnt bandwidths from REPLAY. Figure \ref{week_sigma} shows the values of the learnt bandwidths for the 168 hour-in-week timestamps and the corresponding location prediction performance in MRR for each timestamp on both Gowalla and Foursquare datasets. 

First, we find a clear daily pattern, where the bandwidths in the morning have smaller values in general compared to other time periods; this implies a stronger temporal regularity of the morning mobility (often working-related activities), as the smoothed timestamp embeddings require less information from the neighboring timestamps for location prediction. In contrast, the learnt bandwidths in the nighttime often have larger values, implying a weaker temporal regularity in the nighttime (often entertainment-related activities in LBSNs), as the smoothed timestamp embeddings require more information from the neighboring timestamps for location prediction. Moreover, we further compare the bandwidths between weekdays and weekends, by plotting the average bandwidth for each hour on weekdays and on weekends, as shown in Figure \ref{weekday_sigma} and \ref{weekend_sigma}, respectively. We observe that despite the similarity of having smaller bandwidths in the morning, the bandwidths on weekends have larger values than on weekdays, which implies that weekend activities are less regular than weekdays in general. Moreover, we find that the weekend afternoon sometimes shows the weakest regularity evidenced by the largest learnt bandwidth. 

Second, we also discover a distinct daily pattern in the performance of location prediction. As shown in Figure \ref{all_sigma}, the location prediction performance shows an \textit{inverse} trend compared to the bandwidths. This is due to the fact that the weaker regularity of user mobility at a certain timestamp (in the nighttime or on weekends, for example) makes it harder for location prediction; subsequently, the learnt bandwidth has a larger value, as the query timestamp is less confident for location prediction and thus resorts more to its neighboring timestamps for prediction, and vice versa. 

\begin{table*}[]
\caption{Location prediction performance and the corresponding average bandwidth in different prediction periods. }
\label{day_night}
\centering
\begin{tabular}{c|c|c|c|c}
\hline
\multirow{3}{*}{Prediction time period} & \multicolumn{2}{c|}{Gowalla} & \multicolumn{2}{c}{Foursquare} \\ \cline{2-5}
& Average MRR  & Average Bandwidth & Average MRR & Average Bandwidth \\
\hline \hline
{Daytime (6:00-18:00)}  & 0.2767 & 0.85 & 0.4856 & 1.76 \\
\hline
{Nighttime (18:00-6:00)}  & 0.2362 & 0.88 & 0.4302 & 2.00 \\
\hline \hline
Weekday & 0.2922 & 0.82 & 0.4841 & 1.84 \\
\hline
Weekend & 0.1898 & 0.98 & 0.3758 & 2.00 \\
\hline
\end{tabular}
\end{table*}

Finally, we quantitatively verify our motivating example where the strength of the temporal regularities varies across
typical time periods (daytime/nighttime and weekday/weekend). Table \ref{day_night} shows the average MRR and learnt bandwidths in typical time periods. We see that the daytime prediction performance is 15.0\% better (with 7.5\% lower bandwidth) on average than the nighttime, while the weekday prediction performance is 41.4\% better (with 11.9\% lower bandwidth) on average than the weekend.

\section{Conclusion}
In this paper, motivated by the time-varying temporal regularities of user mobility, we propose REPLAY, a general RNN architecture designed to capture such varying regularities for location prediction. REPLAY seamlessly incorporates smoothed timestamp embeddings with learnable bandwidths into a flashback mechanism. The timestamp-specific learnable bandwidths can automatically learn to adapt to the temporal regularities of different strengths across different timestamps. We conduct a thorough evaluation using two real-world LBSN datasets. Results show that REPLAY outperforms a wide range of state-of-the-art location prediction methods, with 7.7\%-10.5\% improvement over the best-performing baselines. Moreover, we reveal interesting temporal regularity patterns using the learnt bandwidths: 1) morning mobility consistently shows a stronger regularity compared to other time periods; 2) weekend mobility is less regular than weekdays in general, while the weekend afternoon sometimes shows the weakest regularity.

In the future, we plan to further investigate the temporal regularities varying across different user groups and POI categories.

\ifCLASSOPTIONcompsoc
  \section*{Acknowledgments}
\else
  \section*{Acknowledgment}
\fi

This project has received funding from The Science and Technology Development Fund, Macau SAR (0047/2022/A1, 0123/2023/RIA2, 001/2024/SKL), Jiangyin Hi-tech Industrial Development Zone under the Taihu Innovation Scheme (EF2025-00003-SKL-IOTSC), and UIC Research Grant (UICAG100006). This work was performed in part at SICC which is supported by SKL-IOTSC, University of Macau.


\ifCLASSOPTIONcaptionsoff
  \newpage
\fi

\bibliographystyle{IEEEtran}
\bibliography{manuscript}

\begin{IEEEbiography}[{\includegraphics[width=1in,height=1.25in,clip,keepaspectratio]{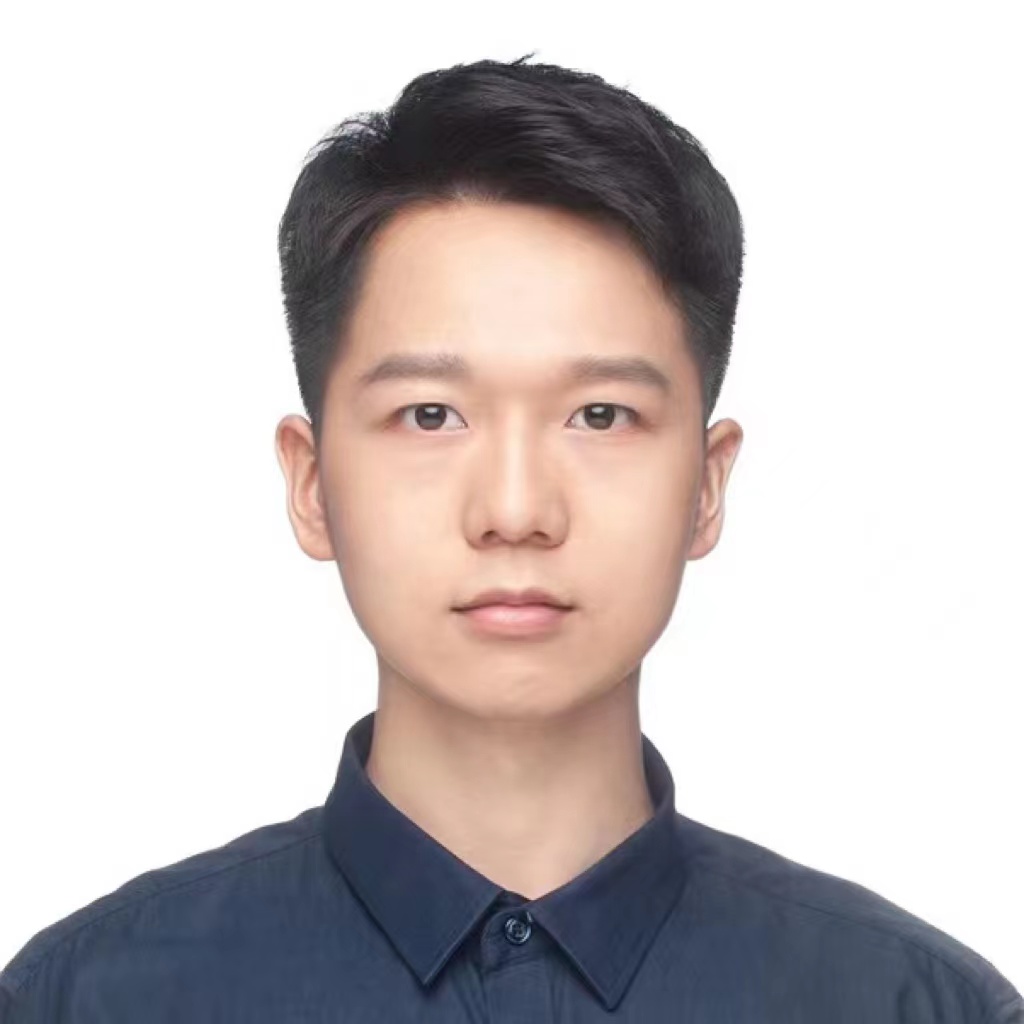}}]{Bangchao Deng} received his B.Eng. degree in Computer Science and Technology from Nanjing University of Aeronautics and Astronautics, China, in 2019. He received his M.S. degree in Computer Science from University of Macau, China, in 2023. He is currently a Ph.D. student with the State Key Laboratory of Internet of Things for Smart City and Department of Computer and Information Science, University of Macau, China. His research interests lie in Spatiotemporal Data Mining and Urban Computing.
\end{IEEEbiography}

\begin{IEEEbiography}[{\includegraphics[width=1in,height=1.25in,clip,keepaspectratio]{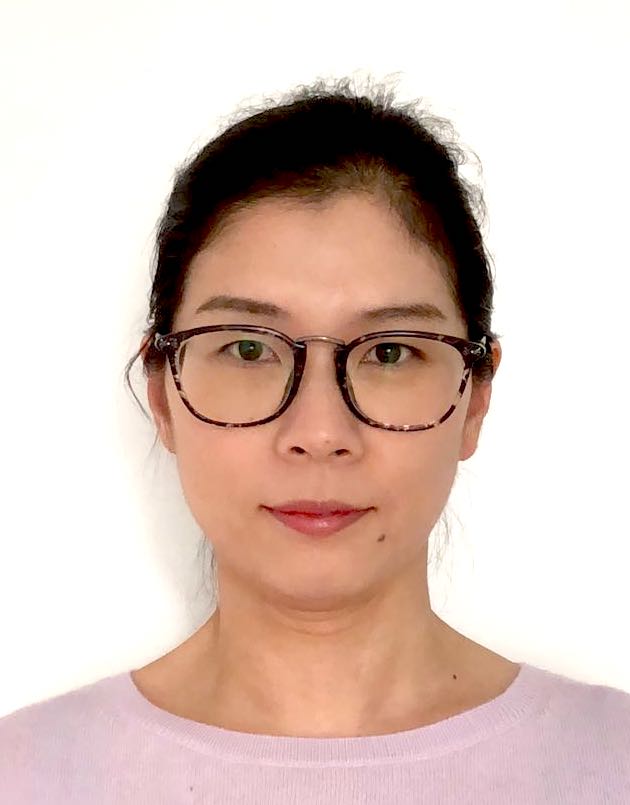}}]{Bingqing Qu}
is an Assistant Professor at the BNU-HKBU United International College. She received her Ph.D. in Computer Science from the University of Rennes 1 in 2016. Before joining the BNU-HKBU United International College, she worked on multimedia data analytics for the Swiss Federal Institute of Technology (EPFL) and the University of Fribourg, Switzerland. Her research interests include multimedia data mining, social media data analytics, and computer vision.
\end{IEEEbiography}

\begin{IEEEbiography}[{\includegraphics[width=1in,height=1.25in,clip,keepaspectratio]{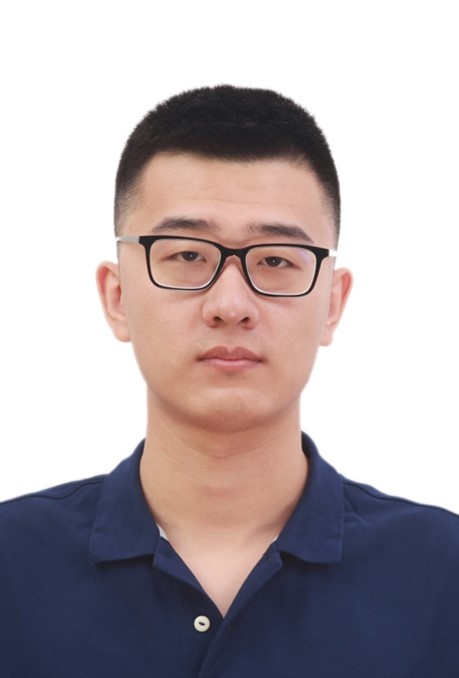}}]{Pengyang Wang}
is an Assistant Professor in the State Key Lab of Smart Cities and Internet-of-Things at the University of Macau. He obtained his Ph.D. in Computer Science from the University of Central Florida. His research interests are in data mining, machine learning and big data analytics. Pengyang has received “Global Top 100 Chinese Rising Stars in Artificial Intelligence”, one Best Student Paper Runner-up award of SIGKDD 2018, and one Best Paper Runner-up award of SIGSPATIAL 2020. His research work has been featured by Synced and UCF Today, and also highlighted by the Natural Science Foundation (NSF) of the U.S.
\end{IEEEbiography}

\begin{IEEEbiography}[{\includegraphics[width=1in,height=1.25in,clip,keepaspectratio]{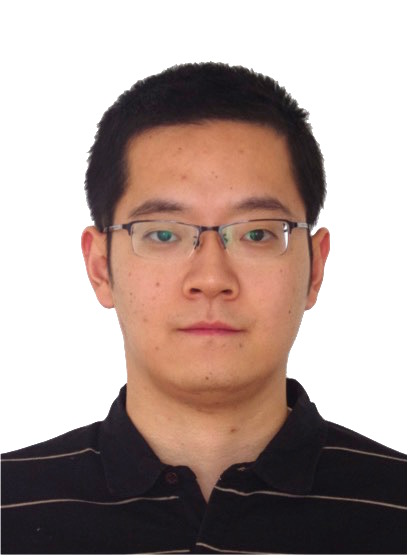}}]{Dingqi Yang}
is an Associate Professor with the State Key Laboratory of Internet of Things for Smart City and Department of Computer and Information Science, University of Macau. He received his Ph.D. degree in Computer Science from Pierre and Marie Curie University and Institut Mines-TELECOM/TELECOM SudParis in France, where he won both the CNRS SAMOVAR Doctorate Award and the Press Mention in 2015. Before joining the University of Macau, he worked as a senior researcher at the University of Fribourg in Switzerland. His research interests include big data analytics, ubiquitous computing, and smart city.
\end{IEEEbiography}

\begin{IEEEbiography}[{\includegraphics[width=1in,height=1.25in,clip,keepaspectratio]{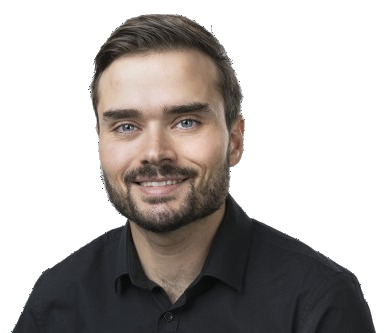}}]{Benjamin Fankhauser}received his Bachelor of Science in Computer Science from the Bern University of Applied Sciences in 2016 and his Master Degree in Computer Science from the University of Bern in 2020. He has worked for various tech companies in Switzerland and is becoming a Ph.D. student at the Pattern Recognition Group at University of Bern. His research interests are in computer vision, data mining and deep learning.
\end{IEEEbiography}

\begin{IEEEbiography}[{\includegraphics[width=1in,height=1.25in,clip,keepaspectratio]{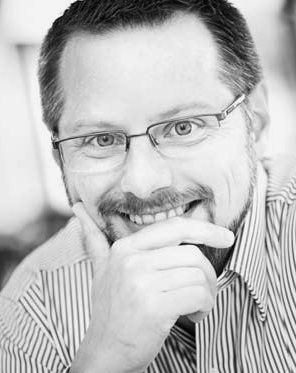}}]{Philippe Cudre-Mauroux}
is a Full Professor and the director of the eXascale Infolab at the University of Fribourg in Switzerland. He received his Ph.D. from the Swiss Federal Institute of Technology EPFL, where he won both the Doctorate Award and the EPFL Press Mention. Before joining the University of Fribourg he worked on information management infrastructures for IBM Watson Research, Microsoft Research Asia, and MIT. His research interests are in next-generation, Big Data management infrastructures for non-relational data. Webpage: http://exascale.info/phil
\end{IEEEbiography}


\end{document}